\documentclass{article}

\usepackage{arxiv}

\usepackage[utf8]{inputenc}  
\usepackage[T1]{fontenc}     
\usepackage{hyperref}        
\usepackage{url}             
\usepackage{booktabs}        
\usepackage{amsfonts}        
\usepackage{nicefrac}        
\usepackage{microtype}       
\usepackage{lipsum}
\usepackage{algorithm}
\usepackage{float}
\usepackage{graphicx}
\usepackage{capt-of}
\usepackage{cuted}
\usepackage{svg} 
\usepackage{cite}
\usepackage{algpseudocode}
\usepackage{amsmath}
\usepackage{amssymb}
\usepackage{multirow}        
\usepackage{tabularx}        
\usepackage{caption}

\title{Temporal convolutional and fusional transformer model with Bi-LSTM encoder-decoder for multi-time-window remaining useful life prediction}

\author{
 Mohamadreza Akbari Pour \\
  Department of Mechanical Engineering\\
  Sharif University of Technology\\
  Teymouri Square, Tarasht, Tehran, Iran\\
  \texttt{mohamadreza.akbari83@sharif.edu} \\
  \And
  Mohamad Sadeq Karimi \\
  Department of Computer Engineering\\
  Sharif University of Technology\\
  P.O. Box 11155-9517 \\
  \texttt{mohamad.karimi@sharif.edu} \\
  \And
 AMIR HOSSEIN MAZLOUMI \\
  Department of Mechanical Engineering\\
  Sharif University of Technology\\
  Teymouri Square, Tarasht, Tehran, Iran \\
  \texttt{mazloumi.amirh@gmail.com} \\
}

\begin{document}
\maketitle
\begin{abstract}
Health prediction is crucial for ensuring reliability, minimizing downtime, and optimizing maintenance in industrial systems. Remaining Useful Life (RUL) prediction is a key component of this process; however, many existing models struggle to capture fine-grained temporal dependencies while dynamically prioritizing critical features across time for robust prognostics. To address these challenges, we propose a novel framework that integrates Temporal Convolutional Networks (TCNs) for localized temporal feature extraction with a modified Temporal Fusion Transformer (TFT) enhanced by Bi-LSTM encoder–decoders. This architecture effectively bridges short- and long-term dependencies while emphasizing salient temporal patterns. Furthermore, the incorporation of a multi-time-window methodology improves adaptability across diverse operating conditions. Extensive evaluations on benchmark dataset demonstrate that the proposed model reduces the average RMSE by up to 5.5\%, underscoring its improved predictive accuracy compared to state-of-the-art methods. By closing critical gaps in current approaches, this framework advances the effectiveness of industrial prognostic systems and highlights the potential of advanced time-series transformers for RUL prediction.

\end{abstract}


\thanks{Preprint Notice: This work has been submitted to \textit{IEEE Access} for possible publication. 
Copyright may be transferred without notice, after which this version may no longer be accessible.}

\section{Introduction}

Prognostics and Health Management (PHM) is a critical discipline focused on predicting the future reliability of systems by assessing their current health state and estimating the Remaining Useful Life (RUL) \cite{FERREIRA2022550}. Accurate RUL estimation is crucial in manufacturing, as it helps prevent unexpected failures, reduce downtime, and lower associated costs. this process relies heavily on condition monitoring data collected from sensors \cite{ABED2024124077} \cite{ZHOU2025125808} \cite{LEE2014314}.\par

The current mainstream RUL prediction methods can be divided into three categories \cite{REZAEIANJOUYBARI2020107929}: physics-based models \cite{8186223}, data-driven methods \cite{FENG2023601}, and their combinations \cite{LIAO2023102195,pour2025accuraterulsohestimation}.
Physics-based models have contributed to PHM but faced practical challenges that limit their use. Though they rely on established principles for predicting system behavior, their dependence on detailed knowledge restricts application in complex or uncertain systems.
Unlike physics-based models, data-driven methods have gained prominence in PHM, primarily due to their adaptability to high-dimensional data environments and their capacity to function without in-depth system knowledge. To achieve accurate predictions of equipment failures, particularly in the context of RUL estimation, these methods leverage advanced statistical techniques, incorporating both machine learning-based and deep learning-based algorithms, to effectively model complex patterns that characterize system behavior \cite{zhao2021deep}. Recently, interest has grown in hybrid methods that combine physics-based and data-driven approaches. A notable example is AttnPINN, which incorporates self-attention within a physics-informed neural network to improve RUL estimation accuracy \cite{LIAO2023102195}. However, the shift towards fully data-driven methods highlights the limitations of hybrid approaches, which often complicate implementation and hinder scalability in addressing modern manufacturing challenges.

Among the foundational deep learning models in PHM are Convolutional Neural Networks (CNNs), known for their exceptional performance in spatial data applications and increasingly applied to time-series fault detection \cite{CNntime}. However, CNNs have limitations in capturing long-range temporal dependencies due to their constrained receptive fields. To address this, Temporal Convolutional Networks (TCNs) have been developed which employ causal and dilated convolutions to model long-term dependencies in sequential data effectively. This makes TCNs particularly suitable for RUL prediction tasks \cite{WANG2021512}.
However, conventional TCNs have a fixed network structure, which limits their flexibility in learning deep temporal representations across varying scales of degradation. This restricts their ability to adapt to complex RUL scenarios where both local fluctuations and long-term degradation patterns must be captured \cite{GAN2024111738}.

CNNs are excellent at organizing spatial data, but superior alternatives exist for sequential data when maintaining the input order is crucial. Recurrent Neural Networks (RNNs) are commonly employed in this field. RNNs are made to handle input data by remembering the most recent inputs, which enforces them with the ability to detect temporal patterns. Although they are still in use, traditional RNNs struggle with long-term dependencies and are limited by the vanishing gradient problem. The Long Short-Term Memory (LSTM) network was developed to overcome this limitation \cite{hochreiter1997}. LSTMs capture long-term dependencies with memory cells and gating mechanisms, making them ideal for RUL prediction in time series where historical data is essential. Building upon this, Bidirectional LSTMs (Bi-LSTMs) process data in both forward and backward directions, which capture context from past and future states.

Even though LSTMs have shown remarkable performance in dealing with temporal relationships, they still have some weaknesses due to their sequential nature, and most of the computing time they use is for parallelization in training. The recent advancements, especially in transformer architecture, are affecting runtime. Unlike RNNs, a transformer is a parallel processing system and does not follow the ordinary way. On the contrary, they simultaneously use an internal attention mechanism to learn the relationship between the tokens from long sequences \cite{vaswani2017attention}. This capability of collecting all time steps simultaneously enables transformers to process more significant sequences using fewer computations \cite{8998569}.
One of the most notable transformer-based architectures for time-series forecasting is the Temporal Fusion Transformer (TFT), which integrates recurrent layers for local processing with interpretable attention layers for long-term dependencies, while also employing variable selection networks and gating mechanisms \cite{lim2021temporal}. TFT has shown strong performance in multi-horizon forecasting tasks with heterogeneous covariates such as retail and healthcare. However, its reliance on static enrichment and single-window processing makes it less suitable for degradation-focused RUL estimation, where the key signals are dynamic sensor measurements and degradation trajectories rather than static metadata.

In recent advancements, hybrid models combining Transformers and LSTMs have been developed to enhance RUL prediction. Fan et al. proposed a Bi-LSTM Autoencoder Transformer model for RUL estimation, to leverage the strengths of both Bi-LSTMs and Transformers to capture intricate temporal patterns in sequential data. In this approach, a Bi-LSTM-based denoising autoencoder is employed to extract robust feature representations from multivariate time-series sensor data, which are then fed into a Transformer encoder for RUL prediction \cite{math11244972}. Another approach involves incorporating channel attention into the model to improve feature extraction for each sensor. The study introduced a multi-feature fusion model that employs CNNs combined with channel attention mechanisms to extract spatial features from sensor data \cite{10063980}.

Despite all the evolutions of deep neural networks and the ongoing adaptations in time series data, relying solely on each of them can result in inaccurate predictions. Furthermore, there are still challenges in integrating deep learning models in PHM, some of which are listed below:

A) The accurate capture of temporal dependencies across varying scales remains a persistent challenge in the field. Still, most models fail to adequately capture both the short and long-term dependencies in the same dataset, which leads to an incomplete understanding of the system.

B) Time-series data can be noisy and high-dimensional, making it difficult for models to disregard channels with misleading information and isolate the most relevant features contributing to system failure. While many methods utilize attention mechanisms to improve prediction, they often fail to dynamically prioritize the most informative patterns within the time series data and might be too complex, which could limit their generalizability to be effectively applied in RUL prediction tasks.

C) Time-series data often contain long sequences of information where early events influence later outcomes. Many models fail to retain critical long-term dependencies, resulting in inaccurate predictions, especially in systems with complex operational histories . 

While architectures such as TFT or TCN provide strong baselines, they remain insufficient for RUL tasks. Standard TFT assumes the availability of static covariates and lacks mechanisms for handling variable-length degradation histories, while a plain TCN lacks gating and long-term memory, leading to an incomplete representation of degradation dynamics. In particular, conventional TCNs rely on fixed receptive fields, which are too rigid to adapt to the evolving multi-scale patterns in engine degradation data \cite{GAN2024111738}. These shortcomings motivate the need for a more flexible architecture that can fuse multi-scale temporal information, dynamically prioritize sensor features, and retain long-term dependencies.

\begin{figure*}[htbp]
	\centering
	\includegraphics[width=0.65\linewidth]{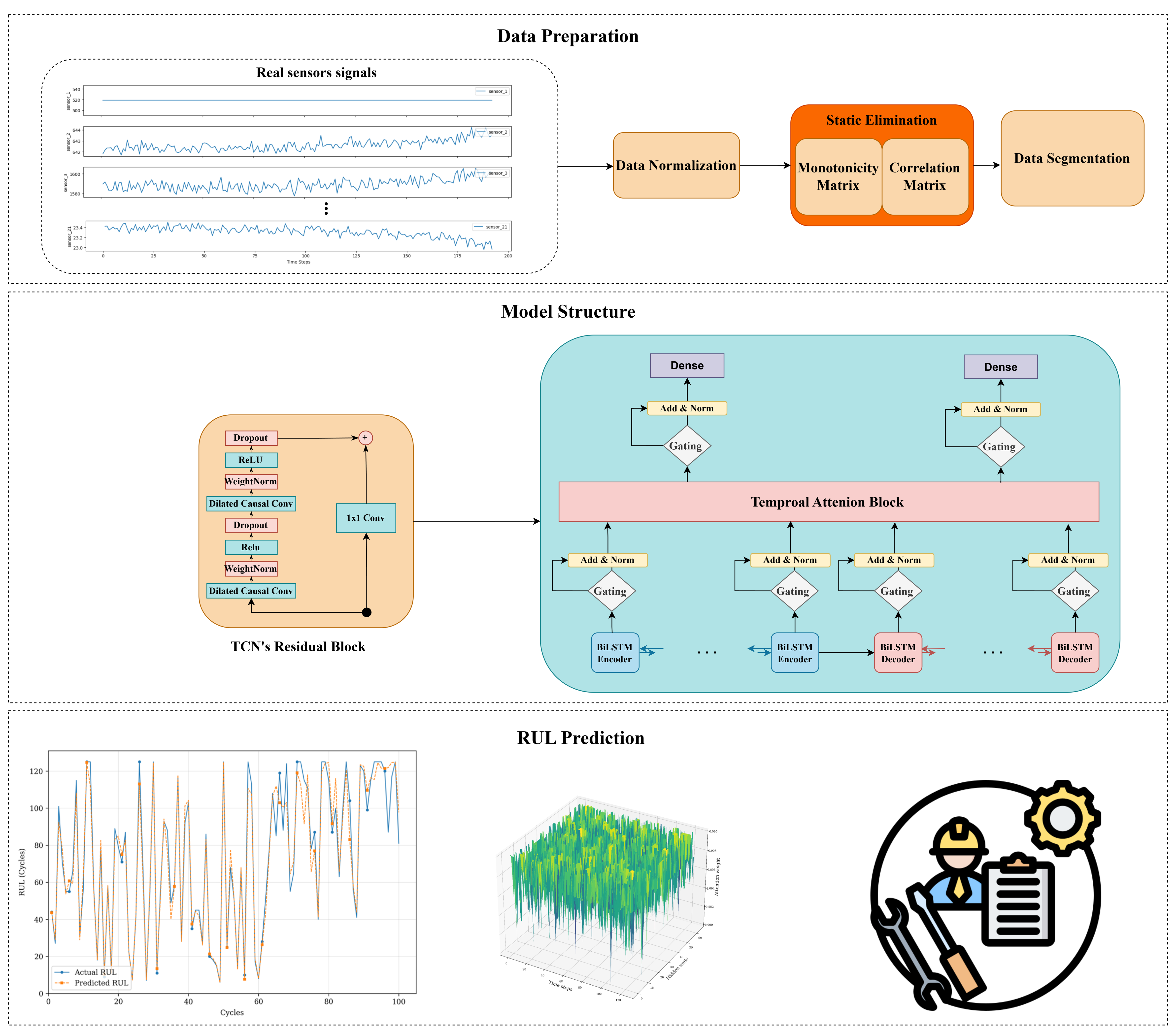}
	\caption{The proposed TCFT-BED's framework.}
	\label{overall}
\end{figure*}

To meet the mentioned challenges, the paper proposes a Temporal convolutional and fusional transformer model with Bi-LSTM Encoder-Decoder for multi-time-window RUL prediction (TCFT-BED) that its comprehensive schematic is presented In Fig. \ref{overall}. The approach employs state-of-the-art methods in time series data and combines adequate layers with different characteristics to capture complex patterns. Altogether, the principal contributions of this study can be summed up as follows:

1. By incorporating TCN, the model processes data across different time scales, allowing it to capture both short-term fluctuations and long-term trends \cite{lea2016temporalconvolutionalnetworksunified}. The multi-time-window technique ensures that temporal patterns at various scales are analyzed and refined, enhancing the model's complexity and, in turn, improving its overall understanding of system behavior. Although the use of TCN introduces higher model complexity and longer training times, it leads to significant improvements in performance. The ablation study shows that the model with TCN achieves a lower RMSE and better score compared to the same model without TCN, demonstrating that the additional complexity results in a more accurate RUL prediction.

2. Unlike the original TFT, which incorporates static covariates (e.g., categorical metadata) into the temporal representation, our modified TFT eliminates the static enrichment layer. While the original TFT, with its static enrichment, introduces more complexity and longer training times, this design is less efficient for RUL tasks, which primarily rely on dynamic sensor signals rather than static context. By removing this layer, our model focuses on the degradation-related dynamics of the sensor data. Irrelevant or static-like features are filtered out using correlation and monotonicity measures before entering the modified TFT block, ensuring that only informative temporal features are processed by the attention and gating mechanisms. This simplification reduces model complexity and improves robustness, leading to better performance. The ablation study confirms that this modification results in a significant RMSE improvement, reducing it from 16.76 to 11.53 on FD001 compared to the standard TFT, making the model not only more efficient but also more accurate for RUL prediction.

3. By leveraging Bi-LSTM layers in an encoder-decoder structure, the model maintains long-term memory and effectively captures temporal dependencies over extended sequences. This is critical when early events in the operational history influence the outcomes. Overall, it provides a robust framework for modeling complex temporal dependencies, generating variable-length outputs, and transforming input features to enhance time series analysis performance.

The rest of the paper is organized as follows: Section 2 introduces the related work on RUL estimation and the structure of the proposed framework. Section 3 details the model's structure. Section 4 analyses the results and evaluates the framework's performance based on the state of the art. Finally, section 5 concludes the paper.

\section{Frameworks}

\subsection{Multi-Time-Window-Based RUL Estimation}
\label{2.1}

\begin{figure}[htbp]
	\centering
	\includegraphics[trim={0 2cm 0 0},clip,width=1\linewidth]{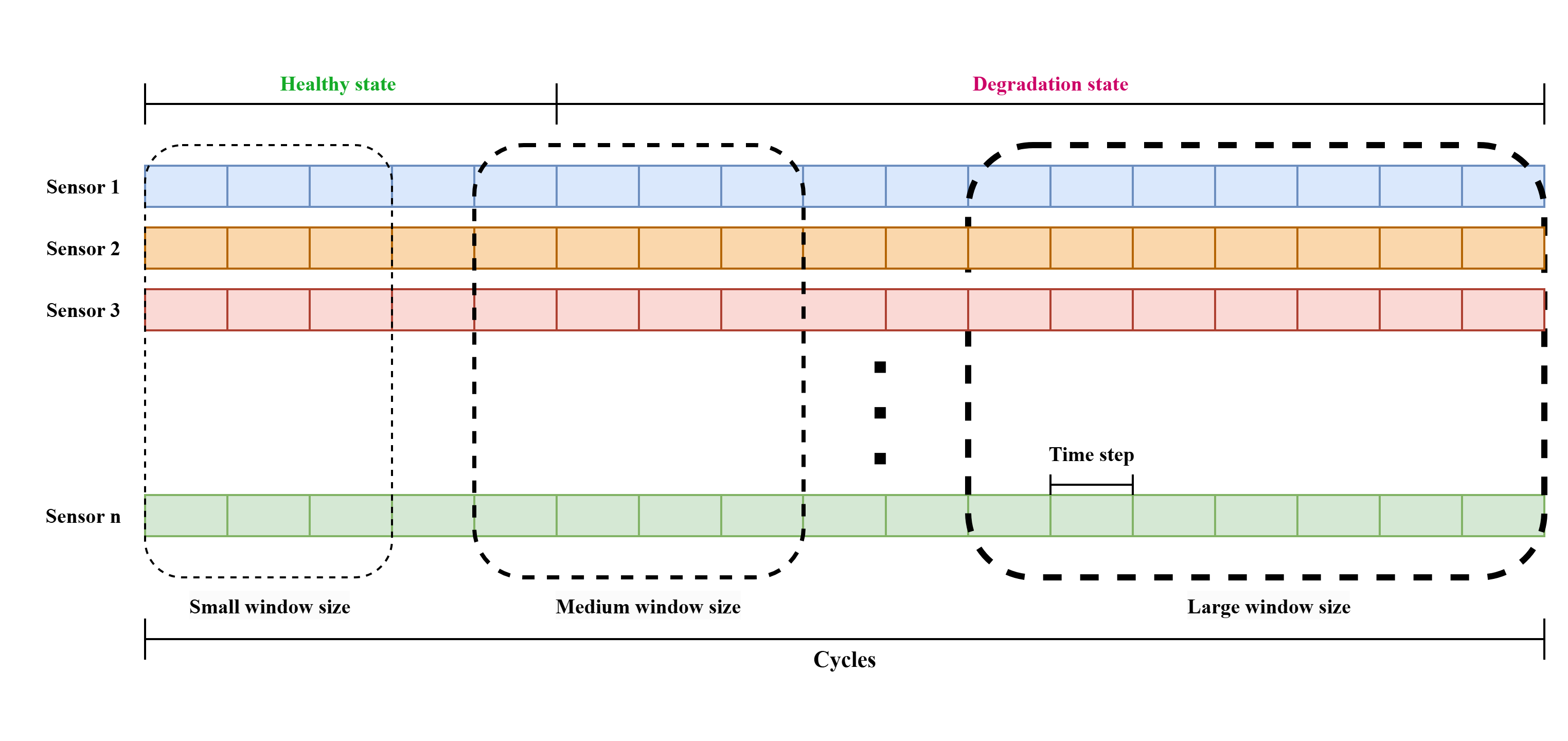}
	\caption{The data segmentation for the RUL prediction.}
	\label{segment12}
\end{figure}
Accurately predicting an engine's behavior and RUL requires monitoring its performance from startup to failure and using data from similar engines under various conditions. How the input data is fed to the model significantly affects its performance. In the case of time series data, using time windows is conventional to handle the sequential nature of the data properly. A static time window approach feeds the model with fixed-length portions where each portion is managed separately. However, it has some limitations, particularly in the case of subtle transitions, because it can't handle engines with different historical data lengths accordingly. To overcome these limitations, a multi-time window approach is proposed. This method applies different time window sizes to capture long- and short-term dependencies. In recent studies, the effectiveness of the method has been highlighted, particularly in applications of hybrid deep neural networks such as civil aircraft auxiliary power unit hazard identification \cite{ZHOU2022344} and turbofan RUL prediction \cite{li2023}. Fig. \ref{segment12},  provides more details about the proposed method in RUL prediction and how the data is segmented.

The model is trained using large, medium, and small window sizes on the entire dataset and evaluated with matching test data for consistent performance assessment. A shift of 1 is used for overlapping windows in the same engine. The multi-time window technique processes test data without overlap between engines to asses the total performance correctly, starting with large time windows, then medium, and finally small, depending on sequence length.


\subsection{Transformers in Time-Series Forecasting: A Paradigm Shift
}
Recent advancements in transformer architectures have significantly transformed the landscape of time-series forecasting, which offers superior performance over traditional methods such as LSTMs and RNNs. Transformers employ attention mechanisms to model temporal dependencies efficiently, which is definitely different from how their predecessors worked, and this is one reason for this kind of application in complex and multi-horizon forecasting.

The TFT stands out as one of the first transformer-based architectures designed specifically for interpretable time-series forecasting. It is recognized for its exceptional performance in the following areas:

\begin{itemize}
	\item \textbf{Variable Selection}: Static and dynamic covariates are prioritized based on their contribution to predictions.
	\item \textbf{Temporal Attention}: Enables the capture of long-term dependencies while focusing on critical time steps.
	\item \textbf{Gating Mechanisms}: Prevents overfitting by regulating the information flow.
\end{itemize}

These features make TFT appealing for applications that need interpretability and robust forecasting. The interpretability of TFT stands out as a pivotal advantage that offers deep insights into the contributions of different features over time. This transparency is invaluable for industrial prognostic systems, where understanding the "why" behind predictions can guide decision-making \cite{lim2021temporal}.

Building upon the foundation laid by TFT, other transformer-based models such as Autoformer and TimesNet have introduced complementary approaches and tackle specific challenges in temporal modeling. While TFT emphasizes dynamic feature selection and interpretable attention mechanisms, Autoformer and TimesNet, mostly focus on distinct aspects of temporal complexity.
Autoformer leverages a decomposition framework to isolate trend and seasonal components, supported by Auto-Correlation Attention for efficient long-sequence dependency modeling. In contrast, TimesNet employs time-convolutional modules and time-aware encoding blocks to capture multi-scale temporal patterns. Altogether, these models highlight the versatility of transformers in tackling diverse temporal complexities \cite{wu2021autoformer, wu2023timesnet}.

\begin{figure*}[htbp]
	\centering
	\includegraphics[width=1\linewidth]{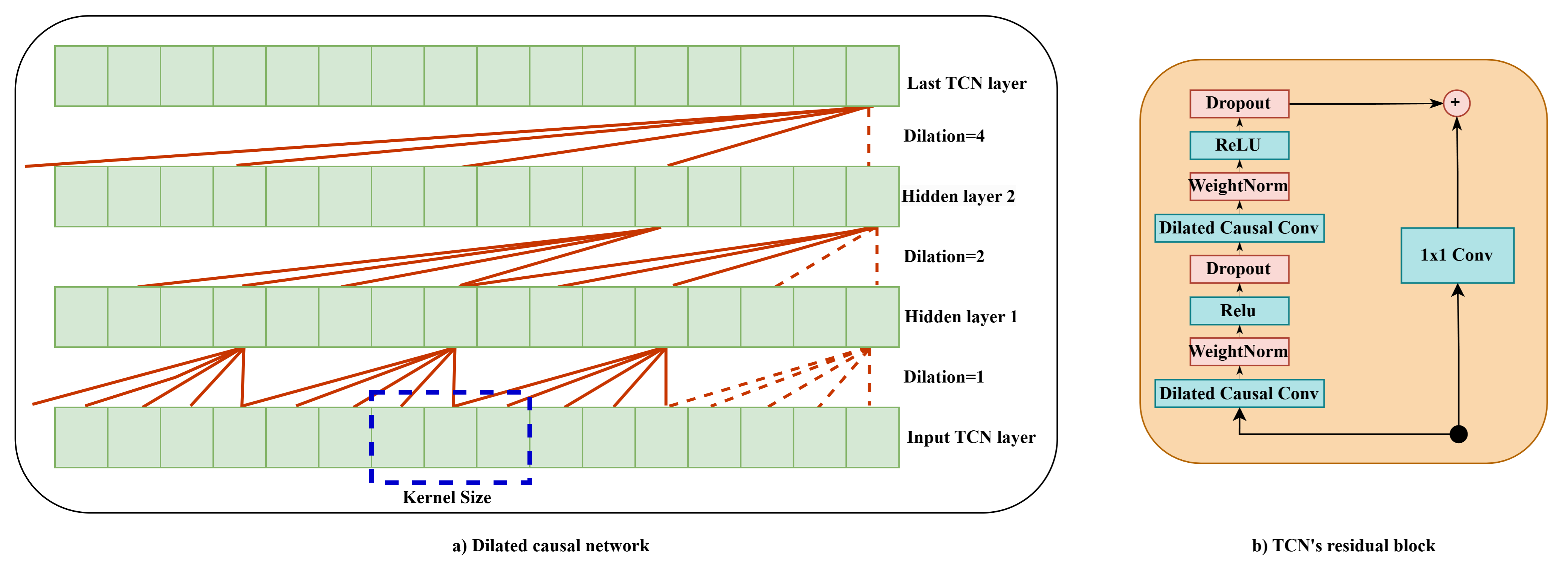}
	\caption{Temporal convolution network's structure.}
	\label{TCNs}
\end{figure*}

\subsection{Temporal Convolutional Networks}
TCNs are increasingly recognized as a viable alternative to RNNs for sequence modeling tasks. In their empirical evaluation, Bai et al. \cite{DBLP:journals/corr/abs-1803-01271} illustrated that TCNs can outperform traditional recurrent models such as LSTM networks and Gated Recurrent Units (GRUs) in various applications. In Fig. \ref{TCNs}, the dilated causal network and TCN's residual are provided in detail. The TCNs' main superiority lies in their ability to model long-range temporal dependencies with considerable computational efficiency due to their unique architectural features.

Causal convolutions ensure the model respects the temporal order of input data, maintaining sequential integrity by preventing predictions at any time step from using future information. Mathematically, a causal convolution for a one-dimensional input sequence can be expressed as:

\begin{equation}
	y(t) = \sum_{i=0}^{k-1} w(i) \cdot x(t - i)
\end{equation}

Where \( y(t) \) represents the output at time \( t \), \( w(i) \) are the learned convolutional filters, and \( x(t - i) \) represents the past inputs up to a receptive field size \( k \). This formulation preserves causality by ensuring the output is dependent solely on past and present inputs.

To further enhance the ability to capture long-range dependencies, TCNs incorporate dilated convolutions, which expand the receptive field by introducing gaps between input values. This allows the model to learn patterns across multiple time scales without significantly increasing the number of parameters. The equation for dilated convolutions is defined as:

\begin{equation}
	y(t) = \sum_{i=0}^{k-1} w(i) \cdot x(t - d \cdot i)
\end{equation}

Where \( d \) is the dilation factor, which increases exponentially with each layer. By expanding the receptive field, TCNs can model longer sequences with fewer layers \cite{oord2016}.

Another critical aspect of TCNs is their incorporation of residual connections. These connections help mitigate the vanishing gradient problem that often affects deep networks by ensuring that information is efficiently passed through the layers. A residual connection can be described as:

\begin{equation}
	z(t) = \text{ReLU}(y(t) + x(t))
\end{equation}

where \( y(t) \) is the output from the convolutional layer, \( x(t) \) is the input to that layer, and \( \text{ReLU} \) represents the Rectified Linear Unit activation function. This structure ensures that information is preserved as it propagates through the network, allowing the model to learn more effectively in deep architectures.

The combination of causal convolutions, dilated convolutions, and residual connections creates a robust architecture that excels in sequence modeling tasks. These features enable TCNs to capture both short- and long-term dependencies in time-series data.

\subsection{Modified TFT with dynamic sensors}
In this section, we discuss the structure of the modified TFT block starting with the LSTM formulations and continuing with the attention block's descriptions and analyzing key aspects of the structure. In the modified TFT, we removed the static enrichment feature to focus on dynamic relations and have a lighter model with less complexity.

\subsubsection{LSTM Formulations}  
The LSTM network is designed to capture long-term dependencies in sequential data through a series of gates that regulate the flow of information. At each time step \( t \), the forward LSTM computes the hidden state \( h_t \) based on the current input \( x_t \), the previous hidden state \( h_{t-1} \), and the previous cell state \( c_{t-1} \). The input gate \( i_t \) controls the amount of new information from the current input \( x_t \) that should be stored in the memory. It is computed as:  
\begin{equation}
i_t = \sigma(W_i x_t + U_i h_{t-1} + b_i),
\end{equation}
with \( \sigma \) being the sigmoid activation function, and \( W_i \), \( U_i \), and \( b_i \) representing the learned weights and bias for the input gate. The forget gate \( f_t \) governs how much of the previous cell state \( c_{t-1} \) is retained. This is expressed as:  

\begin{equation}
f_t = \sigma(W_f x_t + U_f h_{t-1} + b_f),
\end{equation}

where \( W_f \), \( U_f \), and \( b_f \) correspond to the weights and bias for the forget gate. The output gate \( o_t \) determines the portion of the current memory cell that should be output to the next time step, and it is given by:  
\begin{equation}
o_t = \sigma(W_o x_t + U_o h_{t-1} + b_o),
\end{equation}

with \( W_o \), \( U_o \), and \( b_o \) being the weights and bias for the output gate. The candidate memory \( \tilde{c_t} \) represents the potential new information that may be added to the memory, and it is calculated as:  

\begin{equation}
\tilde{c_t} = \tanh(W_c x_t + U_c h_{t-1} + b_c),
\end{equation}

where \( W_c \), \( U_c \), and \( b_c \) are the learned parameters for the candidate memory. Once the gates have been computed, the memory cell \( c_t \) is updated by combining the previous memory \( c_{t-1} \) and the candidate memory \( \tilde{c_t} \), as follows:  

\begin{equation}
c_t = f_t \cdot c_{t-1} + i_t \cdot \tilde{c_t},
\end{equation}
in which the forget gate \( f_t \) determines the proportion of the previous memory to keep, and the input gate \( i_t \) controls the amount of the candidate memory to integrate. Finally, the hidden state \( h_t \) is derived by applying the output gate to the updated memory cell:  

\begin{equation}
h_t = o_t \cdot \tanh(c_t).
\end{equation}
 
This output \( h_t \) either serves as the final output of the LSTM or is fed as input to the subsequent time step in the sequence.

\begin{figure}
	\centering
	\includegraphics[trim={0 1cm 0 0},clip,width=1\linewidth,]{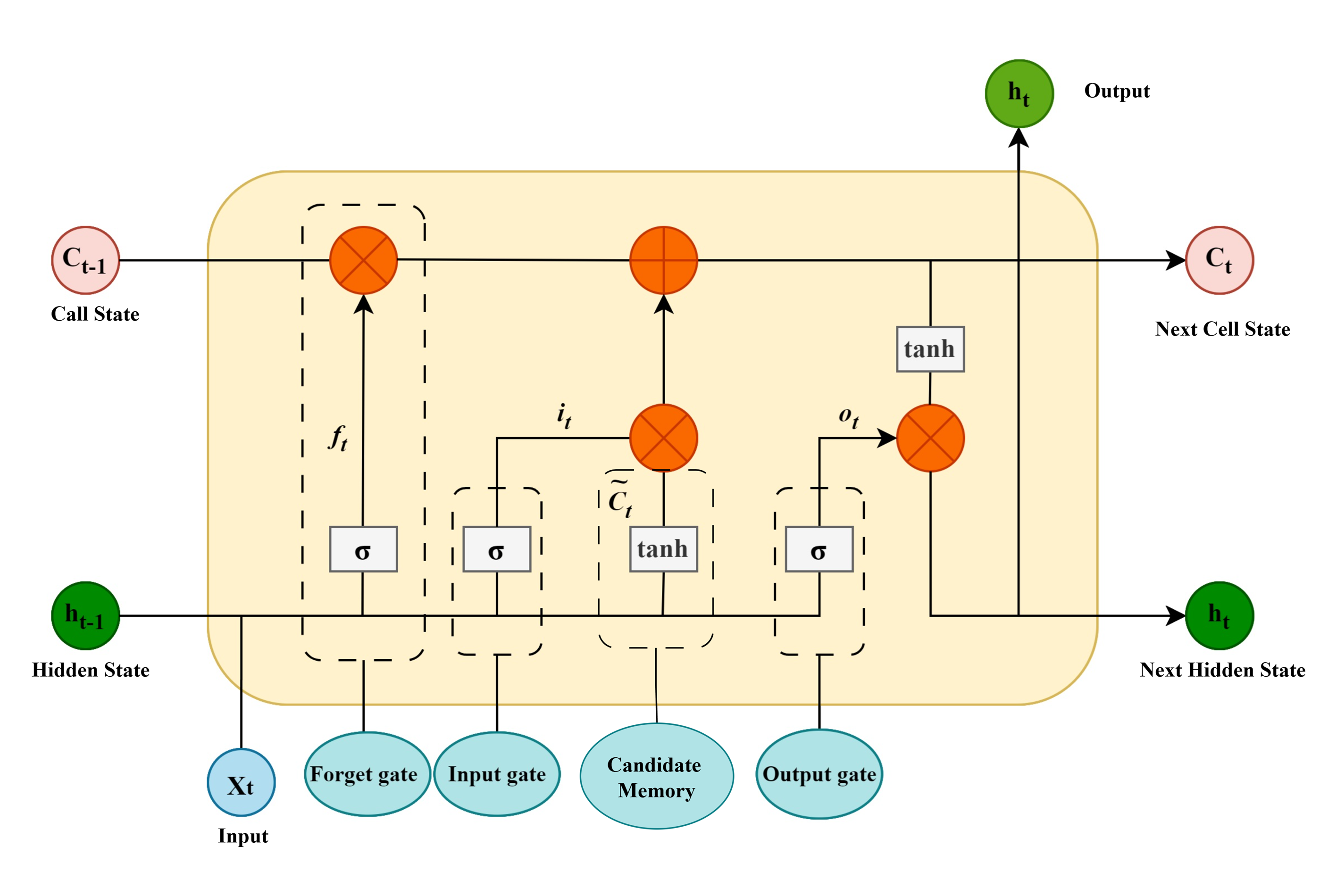}
	\caption{LSTM unit structure.}
	\label{LSTM}
\end{figure}

The updated cell state \( c_t \) and hidden state \( h_t \) are calculated as follows:

\begin{equation}
	c_t = f_t \odot c_{t-1} + i_t \odot \tilde{c_t}
\end{equation}

\begin{equation}
	h_t = o_t \odot \tanh(c_t)
\end{equation}
Here, \( \odot \) denotes the element-wise multiplication (Hadamard product), which allows the model to selectively combine the previous cell state and the candidate memory, as well as to determine the output hidden state based on the current cell state. Fig. \ref{LSTM} provides further details about the LSTM unit structure.

\begin{figure}
	\centering
	\includegraphics[trim={0 0 0 0},clip,width=1\linewidth]{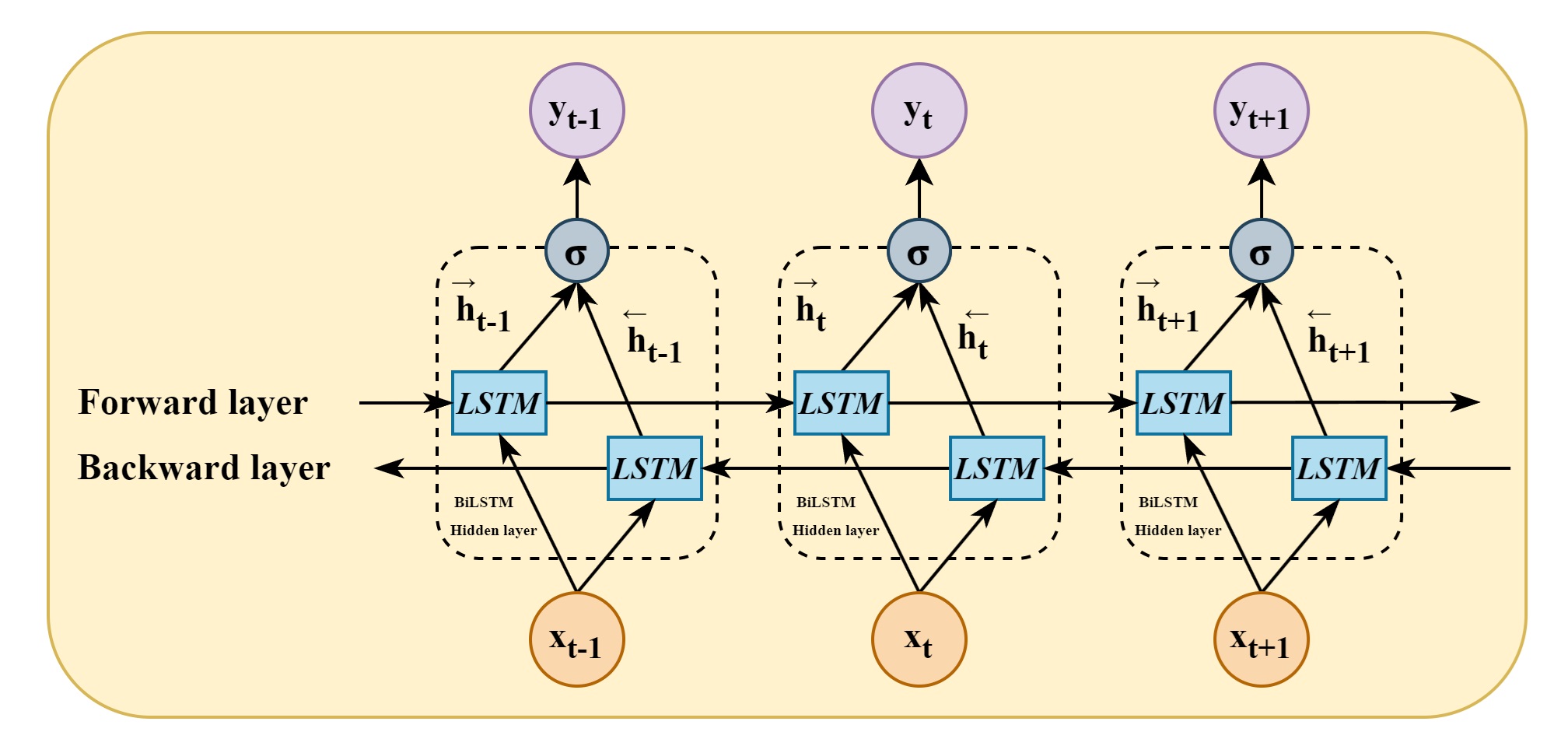}
	\caption{Bi-LSTM network structure.}
	\label{Bi-LSTM}
\end{figure}

\subsubsection{Bi-LSTM formulations}

In the Bi-LSTM, two separate LSTM layers process the sequence in opposite directions. The forward LSTM processes the sequence from \( t = 1 \) to \( T \) and produces forward hidden states \( \{\overrightarrow{h}_t\}_{t=1}^T \). The backward LSTM processes the sequence from \( t = T \) to \( 1 \), yielding backward hidden states \( \{\overleftarrow{h}_t\}_{t=1}^T \).

For a time step \( t \), the hidden states are computed as:
\begin{align}
	\overrightarrow{h}_t &= \text{LSTM}_{\text{forward}}(x_t, \overrightarrow{h}_{t-1}, \overrightarrow{c}_{t-1}), \\
	\overleftarrow{h}_t &= \text{LSTM}_{\text{backward}}(x_t, \overleftarrow{h}_{t+1}, \overleftarrow{c}_{t+1}).
\end{align}

As shown in Fig. \ref{Bi-LSTM}, the final Bi-LSTM hidden state is formed by concatenating the forward and backward hidden states:

\begin{equation}
h_t = \begin{bmatrix}
	\overrightarrow{h}_t \\
	\overleftarrow{h}_t
\end{bmatrix}.
\end{equation}

By combining contextual information from both directions, the Bi-LSTM effectively captures dependencies across the entire sequence.

\subsubsection{Temporal gated attention block}

The attention mechanism in the gated attention block focuses on the most relevant parts of the sequence to learn temporal relationships from the Bi-LSTM encoder's output. The Multi-Head Attention (MHA) mechanism is utilized to compute the attention across different heads in parallel. Formally, the attention mechanism computes:

\begin{equation}
	\text{Attention}(Q, K, V) = \text{softmax} \left( \frac{Q K^\top}{\sqrt{d_k}} \right) V
\end{equation}
Where \( Q \), \( K \), and \( V \) are the query, key, and value matrices, respectively, and \( d_k \) represents the dimensionality of the keys. Multi-head attention applies multiple attention mechanisms in parallel, and the outputs from different heads are concatenated to produce a refined representation of the time steps \cite{vaswani2017attention}. The temporal attention block employs this mechanism to effectively capture temporal relationships within the sequence.

After applying the attention mechanism, a feed-forward layer further refines the representation by linearly transforming the output \cite{ref_ffn_performance}. The feed-forward operation is defined as:

\begin{equation}
	\text{FF}(x) = W_{fc} x + b_{fc}
\end{equation}
Where \( W_{fc} \) is the weight matrix, and \( b_{fc} \) is the bias term in the fully connected layer.

A key addition in this model is the use of gating mechanisms in conjunction with the attention mechanism, in the modified TFT's architecture. Gating mechanisms allow the model to pass relevant information while filtering out less useful parts selectively. This gating operation uses a sigmoid activation to control the flow of information. Formally, the gated output is computed as:

\begin{equation}
	\text{Gated Output}(x) = \sigma(W_g x) \odot x
\end{equation}
Where \( W_g \) is the learned weight matrix, \( \sigma \) denotes the sigmoid activation function, and \( \odot \) represents element-wise multiplication.

\begin{figure}[htbp]
	\centering
	\includegraphics[trim={0 1cm 0 0},clip,height=0.4\textheight]{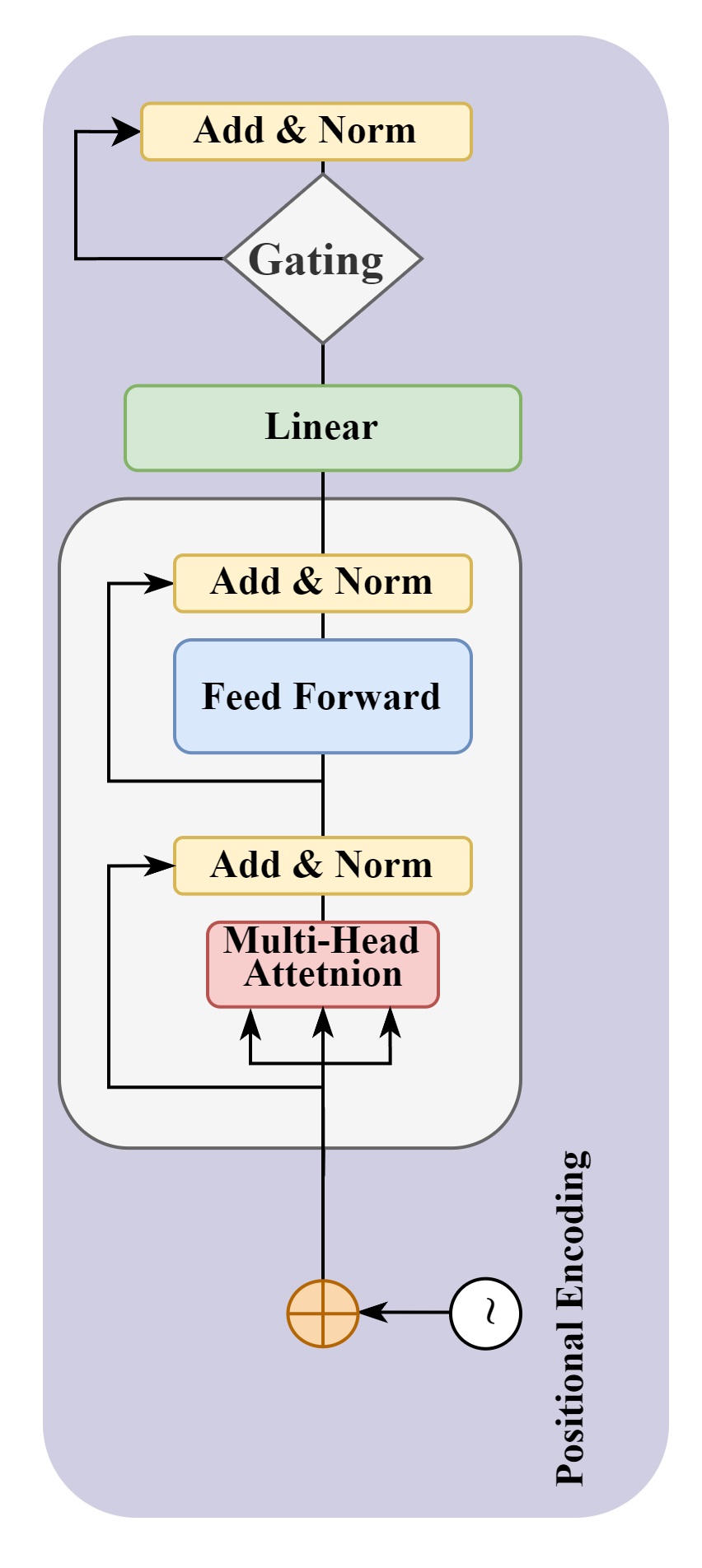}
	\caption{Gated attention block.}
	\label{GTB}
\end{figure}

\begin{figure}[htbp]
	\centering
	\includegraphics[trim={0 1cm 0 0},clip,width=1\linewidth]{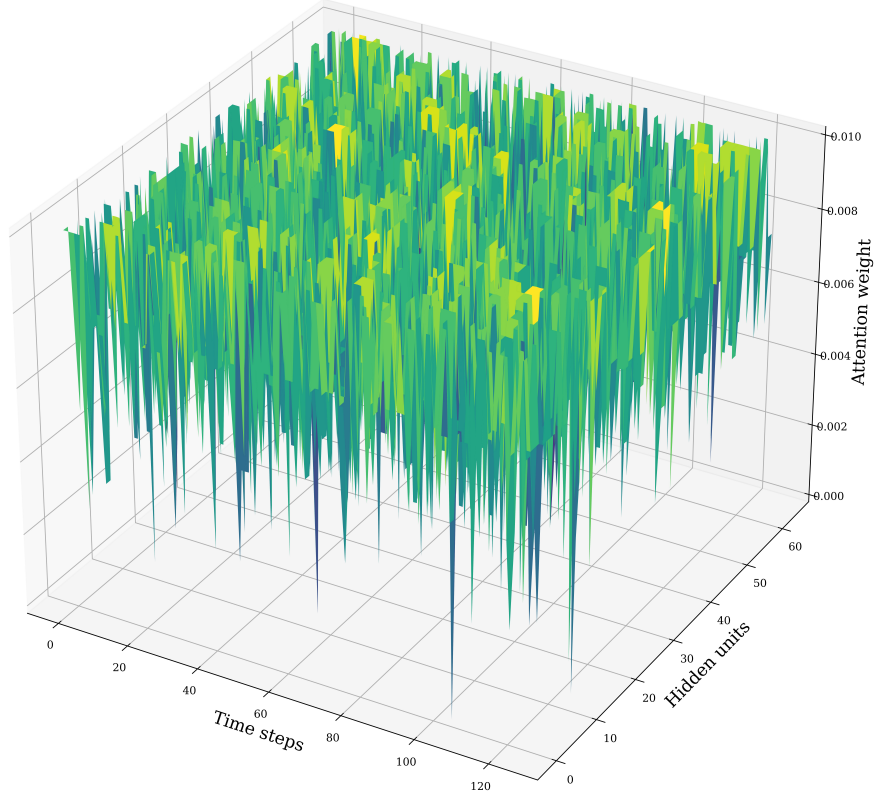}
	\caption{Three-dimensional plot of attention weights for a random testing engine.}
	\label{3dweight}
\end{figure}

To ensure stability during training, the output of the attention block is normalized using layer normalization. This step helps stabilize and accelerate the training process \cite{Ba2016}. As shown in Fig. \ref{GTB}, the gated attention block combines the gating mechanism with multi-head attention. The visualization in Fig. \ref{3dweight}, highlights the dynamic attention weights' distribution across both time steps and hidden units.

\begin{figure*}[htbp]
	\centering
	\includegraphics[width=0.75\linewidth]{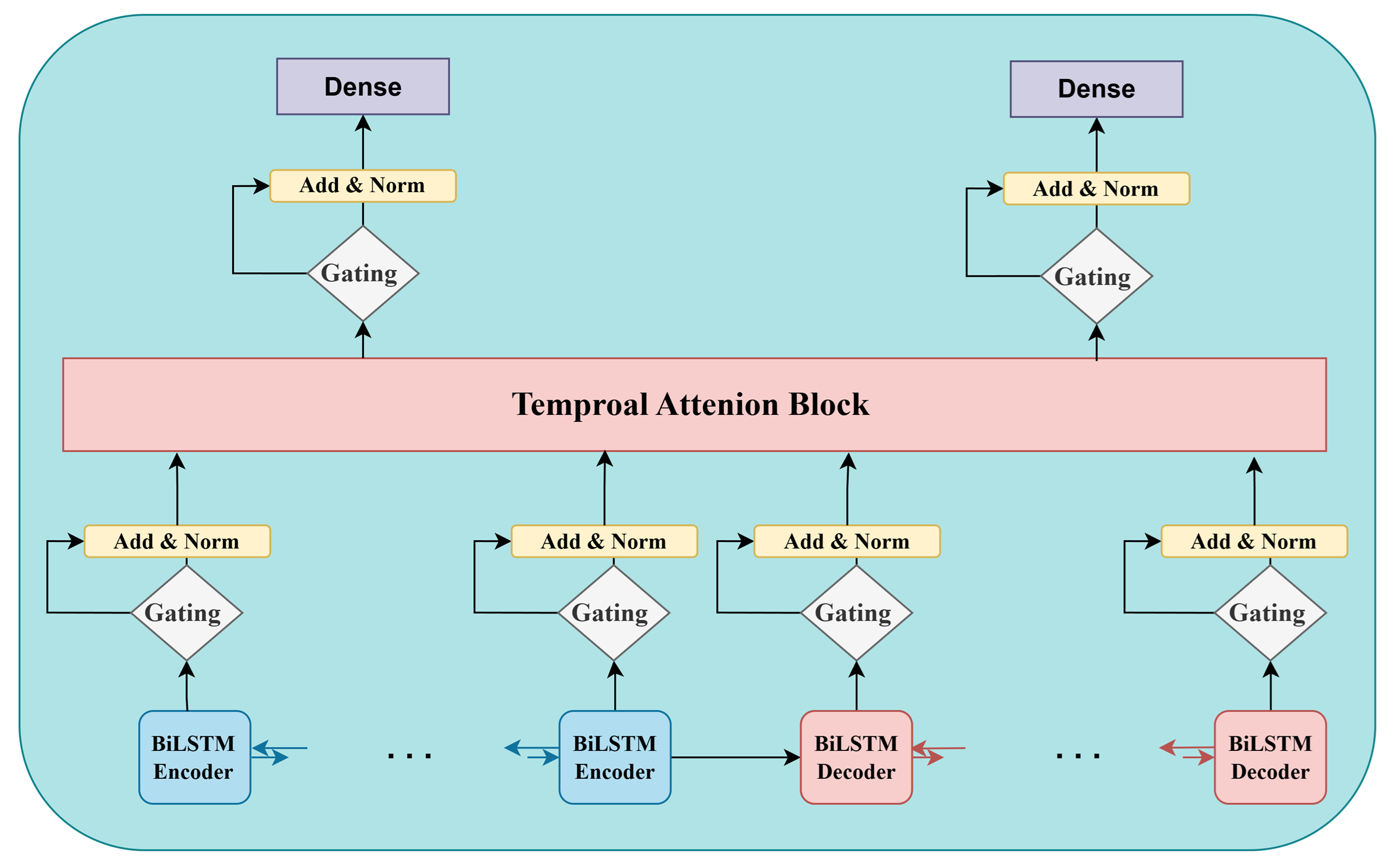}
	\caption{Modified temporal fusion transformer framework.}
	\label{encoder}
\end{figure*}

After passing the data through the gating mechanism, it is further processed by the decoder component, which enhances the model’s ability to capture temporal dependencies. The output of the attention mechanism is then passed through a Bi-LSTM decoder. This step reinforces sequential continuity in the degradation trajectories, ensuring that the globally fused features from attention are reconstructed into a temporally coherent representation suitable for RUL prediction. As illustrated in Fig. \ref{encoder}, the modified Temporal Fusion Transformer (TFT) omits the static enrichment component. In the original TFT architecture, static enrichment layers and static covariate encoders were introduced to condition temporal features on categorical or time-invariant metadata (e.g., entity IDs, location, or demographic attributes) \cite{lim2021temporal}. However, in RUL prediction tasks—such as those involving the C-MAPSS dataset—such static covariates are not usually provided or informative. The degradation behavior is predominantly governed by multivariate sensor signals. As a result, we remove the static enrichment layers to reduce unnecessary architectural complexity and allow the modified TFT to focus entirely on dynamic sensor inputs.

This design choice is further supported by our feature selection framework, which filters out irrelevant and static sensor variables using correlation and monotonicity-based metrics. By retaining only the most informative features, the model is able to eliminate both static covariates and uninformative dynamic inputs. This enables the architecture to more effectively capture critical temporal patterns essential for accurate RUL prediction.

Table 1 provides a concise summary of the proposed algorithm, outlining its main components and the corresponding steps involved in our method.

\begin{table}[htbp]
	\centering
	\caption{
		Algorithm summary.}
	\begin{tabular}{p{0.9\linewidth}}
		\hline
		\textbf{Input:} 3D tensor X(batch size, channels, window size) \\ 
		\textbf{Output:} Trained TCFT-BED model for RUL prediction \\ \hline
		
		\textbf{Data preprocessing:} \\
		Standardizing input data, Sensor selection, Applying multi-time window approach, Converting to PyTorch tensors, Creating data loader objects\\
		
		\textbf{Model initialization:} \\
		Instantiate TCFT-BED model by applying Xavier uniform initialization. \\ 
		
		\textbf{Training:}  \\
		For each epoch: \\
		\hspace{1em} \textbullet\ Pass the input through the TCN block and utilize (1-3) to process it. \\
		\hspace{1em} \textbullet\ Feed features to the Bi-LSTM encoder through (3-11); obtaining encoded temporal features. \\
		\hspace{1em} \textbullet\ Pass Bi-LSTM encoder outputs to the next gating mechanism and then to the temporal gated attention block for attention-based feature integration with (12-14); Obtaining detailed attention on the the most informative patterns within the sequences.  \\
		\hspace{1em} \textbullet\ Feed the attention-based features to the Bi-LSTM decoder layer for further temporal processing by using (3-11) and finally feed outputs to the dense layers. \\
		\hspace{1em} \textbullet\ Compute loss using Mean Squared Error (MSE).\\
		\hspace{1em} \textbullet\ Backpropagate errors and update model parameters using Adam optimizer. \\ \hline
	\end{tabular}
	\label{table:algorithm_summary}
\end{table}

\section{Experimental testing}

In this section, the paper provides experimental test results on C-MAPSS dataset and evaluates TCFT-BED's performance. Finally, the proposed model's superiority compared to other models is demonstrated, and the model analyses are provided.

\subsection{Dataset}
C-MAPSS dataset contains historical data of turbofan engines that operated in different operational conditions and stopped working due to different fault modes. Each engine operates at the start of its time series and develops a fault at some point. The dataset comprises four subsets, each containing separate training and testing data: FD001, FD002, FD003, and FD004. It is organized in a structured tabular format with 26 columns, representing various features relevant to the prediction task. Column 1 is the engine unit number, and column 2 provides the cycle number over time. Columns 3-5 are operational settings, and the remaining 21 columns correspond to sensor measurements. Each sensor measures specific physical properties of the data, such as speed, temperature, and pressure, and is located in a particular part of a turbofan engine \cite{4711414}.

Table~\ref{C-MAPSS_datasets} summarizes the key properties of each C-MAPSS subset, including the number of train/test trajectories, operating conditions, and fault modes. In this dataset, the training and test sets are predefined and provided separately, eliminating the need for manual data splitting. FD001 is the simplest subset, with one operating condition and one fault mode. FD002 introduces higher complexity by including six different operating conditions while retaining a single fault mode. FD003 includes two fault modes but operates under a single condition. FD004 is the most complex, combining six operating conditions with two fault modes. The increased variability in FD002 and FD004 makes them more challenging, as models must learn to generalize across diverse flight scenarios. Overall, the subsets represent a clear progression in difficulty, offering a scalable benchmark for evaluating prognostics models.

\begin{table*}[htbp]
    \centering
    \small
    \setlength{\tabcolsep}{4pt} 
    \renewcommand{\arraystretch}{1.2}
    \caption{C-MAPSS dataset: operating conditions, fault modes, and trajectory counts.}
    \label{C-MAPSS_datasets}
    \begin{tabular}{c c c c c}
        \toprule
        Dataset & Train Trajectories & Test Trajectories & Conditions & Fault Modes \\
        \midrule
        FD001 & 100 & 100 & 1 (Sea Level) & 1 (HPC Degradation) \\
        FD002 & 260 & 259 & 6 & 1 (HPC Degradation) \\
        FD003 & 100 & 100 & 1 (Sea Level) & 2 (HPC Degradation, Fan Degradation) \\
        FD004 & 248 & 249 & 6 & 2 (HPC Degradation, Fan Degradation) \\
        \bottomrule
    \end{tabular}
\end{table*}

\subsection{Data Preparation}

In data-driven approaches, adequate data preparation is critical to model success. This process begins with cleaning and normalizing raw data to ensure consistent feature scaling. Monotonicity and correlation matrices are then used to evaluate relationships between variables. They are mainly useful for feature selection and improving model robustness, which is helpful for data-driven approaches. Through comprehensive data preparation, models with better accuracy, reduction of the rejects, and capability to generalize better across the different situations were obtained.

\subsubsection{Min-max normalization}

The sensor data is normalized using min-max normalization to scale the values to a range of [0, 1], ensuring uniform scaling across sensors. Min-max normalization is calculated as follows:

\begin{equation}
x' = \frac{x - x_{\text{min}}}{x_{\text{max}} - x_{\text{min}}}
\end{equation}

Where:
\begin{itemize}
	\item $x$ is the original sensor value,
	\item $x_{\text{min}}$ and $x_{\text{max}}$ are the minimum and maximum values of the sensor readings, respectively.
\end{itemize}

This ensures that all sensor data is scaled consistently before being input into the model.

\subsubsection{Static Elimination}

Monotonicity and correlation matrices are two preferred methods for evaluating sensor data trends and their relationship with time cycles. The paper employs them to successfully select the most critical sensors and use them in the training and testing. Monotonicity measures how consistently a sensor's readings increase or decrease over time, whereas the correlation matrix captures the strength of the linear relationship between sensor values and time cycles. These matrices guide feature selection and help retain sensors exhibiting strong correlations and consistent trends. The correlation matrix for each sensor with the time cycle can be calculated using the Pearson correlation coefficient:

\begin{equation}
r = \frac{\sum_{i=1}^{n} (x_i - \bar{x})(y_i - \bar{y})}{\sqrt{\sum_{i=1}^{n} (x_i - \bar{x})^2} \sqrt{\sum_{i=1}^{n} (y_i - \bar{y})^2}}
\end{equation}

Where:
\begin{itemize}
	\item $x_i$ and $y_i$ are the values of the sensor and time cycles at the $i^{th}$ observation,
	\item $\bar{x}$ and $\bar{y}$ are the mean values of the sensor readings and time cycles, respectively.
\end{itemize}

\begin{figure*}[htbp]
	\centering
	\includegraphics[width=1\linewidth]{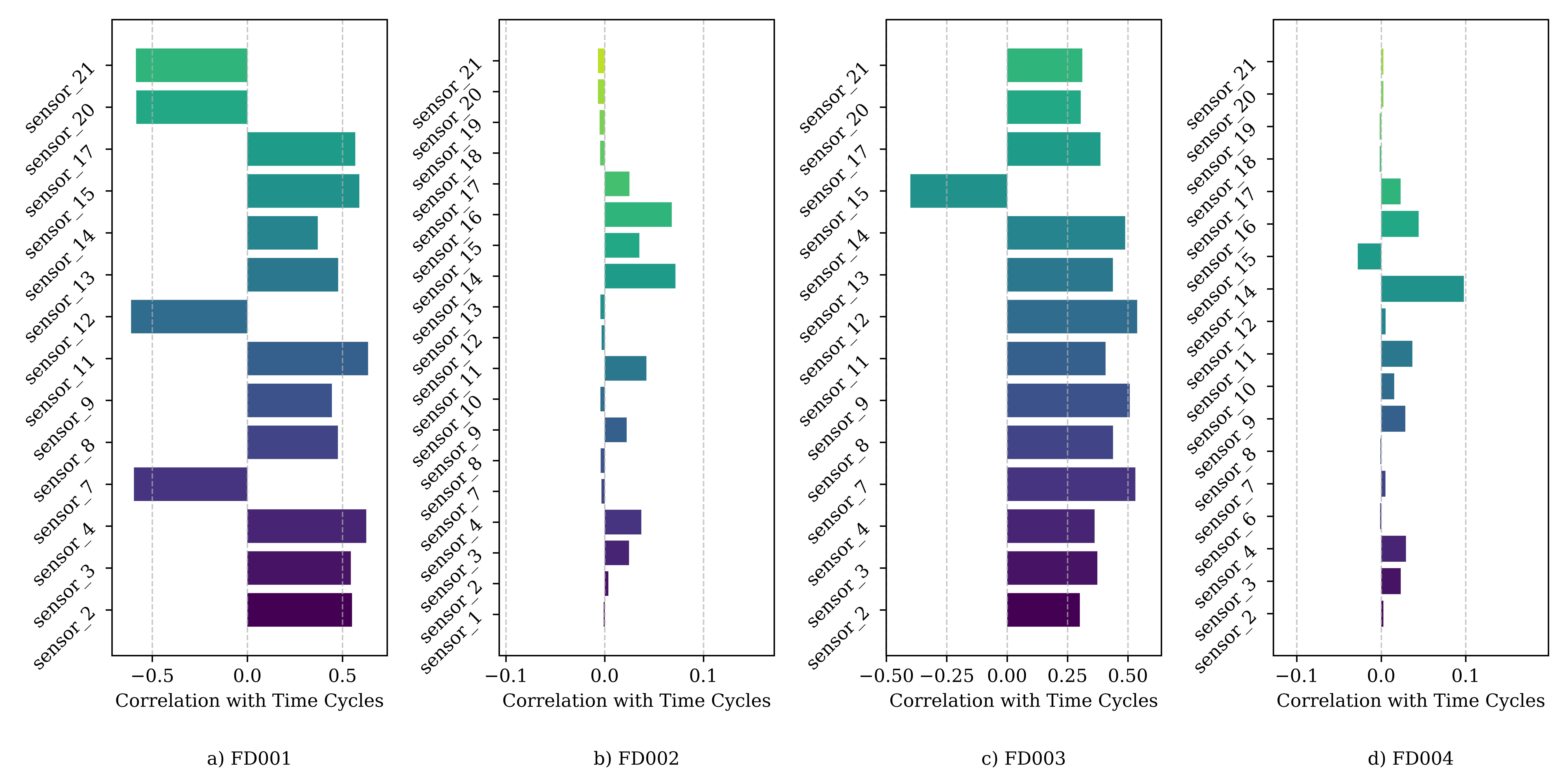}
	\caption{Selected sensors and their correlation with time cycles for each sub-dataset.}
	\label{Selected_sensor}
\end{figure*}

The monotonicity of each sensor is calculated by assessing the proportion of time the sensor values either increase or decrease consistently. This is defined as:

\begin{equation}
M = \frac{1}{n-1} \sum_{i=1}^{n-1} \mathbb{1}(x_{i+1} - x_i > 0)
\end{equation}
Where $\mathbb{1}$ is an indicator function that returns 1 if the condition is true and 0 otherwise, based on these calculations, sensors are selected if their absolute correlation exceeds a threshold and their monotonicity is sufficiently high. Mathematically, this selection process can be written as:

\begin{equation}
S = \{x : |r(x, y)| > \tau_c \ \text{and} \ M(x) > \tau_m \}
\end{equation}
Where $\tau_c$ and $\tau_m$ are predefined thresholds for correlation and monotonicity, respectively. The proposed method guarantees that only sensors with solid correlations to time cycles and consistent increasing or decreasing trends are used for model training. For more visualization, Fig. \ref{Selected_sensor} presents the correlation value of sensors after selection with time cycles that illustrate FD001 and FD003 sensors have a higher correlation to time cycles and fewer selected sensors than FD002 and FD004.

\subsubsection{Piece-wise RUL}
\begin{figure}[htbp]
	\centering
	\includegraphics[width=0.75\linewidth]{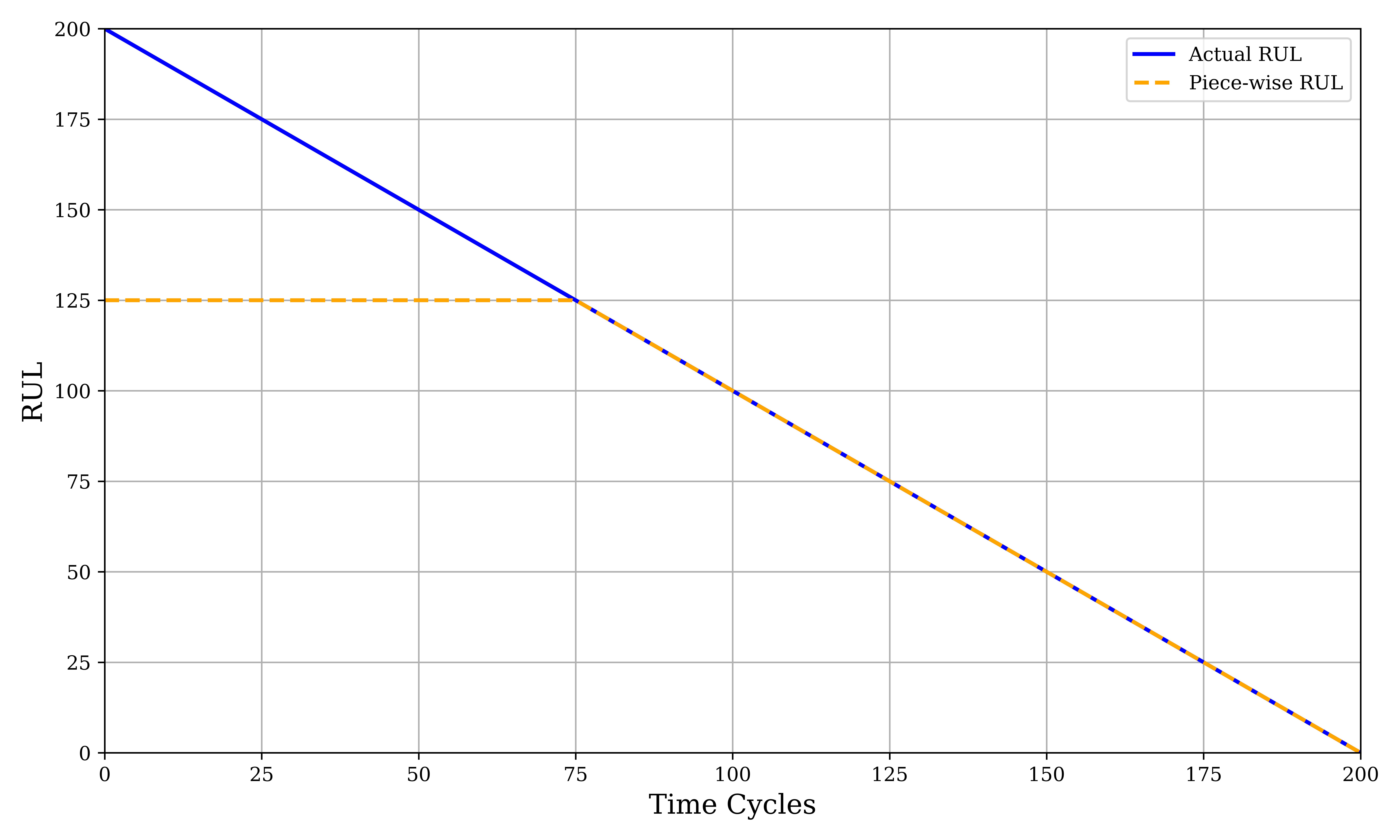}
	\caption{Piece-wise RUL of C-MAPSS dataset, 
		maximum RUL is set at 125 time cycles. }
	\label{Piecewise}
\end{figure}

Run-to-failure sequential data is common in RUL estimation. Usually, the lifetime of an engine can be divided into a healthy state and a degradation state that starts after a fault occurs. The model assumes that the degradation is linear after the initial fault. This study utilizes the piece-wise function, and a similar maximum RUL of 125 is set for every engine. Fig. \ref{Piecewise} provides more details about the piece-wise function \cite{10251604}.

\begin{table*}[htbp]
    \caption{Model components and parameters.}
    \centering
    \label{hyper}
    \renewcommand{\arraystretch}{1.2}
    \begin{tabularx}{\textwidth}{XXXX}
        \toprule
        \textbf{Component} & \textbf{Description} & \textbf{Parameter} & \textbf{Value} \\ 
        \midrule
        
        \multirow{2}{*}{TCN Block} 
            & \multirow{2}{=}{Temporal pattern extraction using dilated convolutions} 
            & Number of filters & 64 \\
            & & Kernel size / Dilation & 3 / \{1,2,4,8,16\} \\
        
        LSTM Encoder 
            & Encodes sequential patterns from TCN output 
            & Hidden dim / Bidirectional & 64 / Yes \\

        \multirow{3}{*}{Modified TFT Block} 
            & \multirow{3}{=}{Attention and gating for enhanced feature learning} 
            & Hidden dimension & 128 \\
            & & Number of heads & 8 \\
            & & Dropout rate & 0.2 \\
        
        LSTM Decoder 
            & Refines sequential predictions 
            & Hidden dim / Bidirectional & 64 / Yes \\

        Fully Connected Layers 
            & Maps LSTM output to final predictions 
            & FC1 / FC2 outputs & 64 / 1 \\

        Dropout & Regularization & Dropout rate & 0.2 \\

        Weight Initialization & Initialization scheme & Type & Xavier uniform \\

        \multirow{2}{*}{Optimizer} 
            & \multirow{2}{=}{Adaptive gradient optimization} 
            & Type & Adam \\
            & & LR / Weight decay & 0.001 / 1e-5 \\

        Loss Function & Error measure & Type & MSE \\

        Training Epochs & Number of epochs & Value & 100 \\

        Evaluation Metric & Performance metrics & Type & RMSE, Score \\ 
        
        \bottomrule
    \end{tabularx}
\end{table*}

\subsubsection{Performance Metrics}

In this study, to evaluate the model's performance and compare it with similar studies on the same dataset, we used the RMSE and score as the performance metrics.

\textbf{Root Mean Squared Error (RMSE):} RMSE measures the square root of the average squared differences between predicted and true RUL values. It is sensitive to significant errors and is commonly used for regression tasks. The RMSE is computed as:

\begin{equation}
\text{RMSE} = \sqrt{\frac{1}{n} \sum_{i=1}^{n} (y_i - \hat{y}_i)^2}
\end{equation}
Where $y_i$ are the true RUL values, $\hat{y}_i$ are the predicted RUL values, and $n$ is the number of observations.

\textbf{Score:} The score penalizes late predictions more heavily than early ones, emphasizing the importance of timely maintenance predictions. It is defined as:

\begin{equation}
S(\hat{y}, y) =
\begin{cases} 
	e^{-\frac{\hat{y} - y}{13}} - 1 & \text{if} \ \hat{y} \geq y \\
	e^{\frac{y - \hat{y}}{10}} - 1 & \text{if} \ \hat{y} < y
\end{cases}
\end{equation}

\begin{figure}
	\centering
	\includegraphics[width=0.75\linewidth]{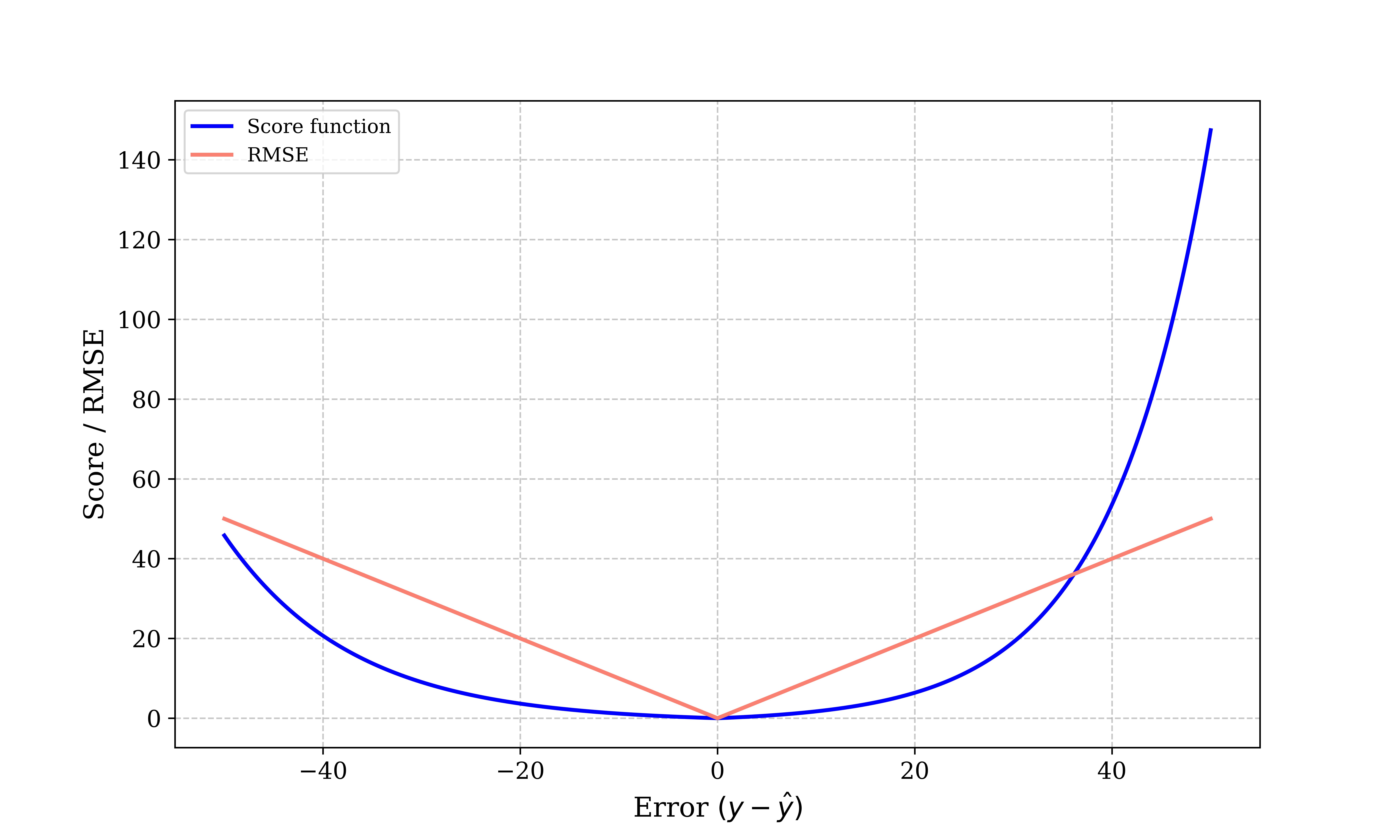}
	\caption{Score and RMSE comparison.}
	\label{SoreVSRMSE}
\end{figure}

Where $\hat{y}$ and $y$ represent the predicted and true RUL values, respectively. Fig. \ref{SoreVSRMSE}, compares the metrics and illustrates score's sensitivity to higher errors. 

	\subsection{Experimental Setup and Implementation Details}
	
	This section will discuss the selection of model parameters used in developing the proposed TCFT-BED. The parameters were chosen based on the requirements to balance model complexity, generalization capability, and efficient training. Table \ref{hyper} provides the exact details about the model's hyper-parameters, where the TCN block was configured with a moderate number of filters and dilation to efficiently capture short- and long-term dependencies in time-series data. In the next part, we employed Bi-LSTMs as encoder-decoder layers for their proven ability to model complex temporal dependencies in both forward and backward directions. After the encoder, the temporal attention block was introduced for enhanced feature fusion and attention, utilizing multi-head attention to focus on critical temporal information, and the second Bi-LSTM was employed to decode the output of the temporal attention block. Finally, the Adam optimizer was selected for its efficiency in handling large, sparse gradients, with regularization applied via weight decay and dropout.

	\subsection{Multi-Time-Window Approach}
	
\begin{figure}
		\centering
		\includegraphics[width=0.75\linewidth]{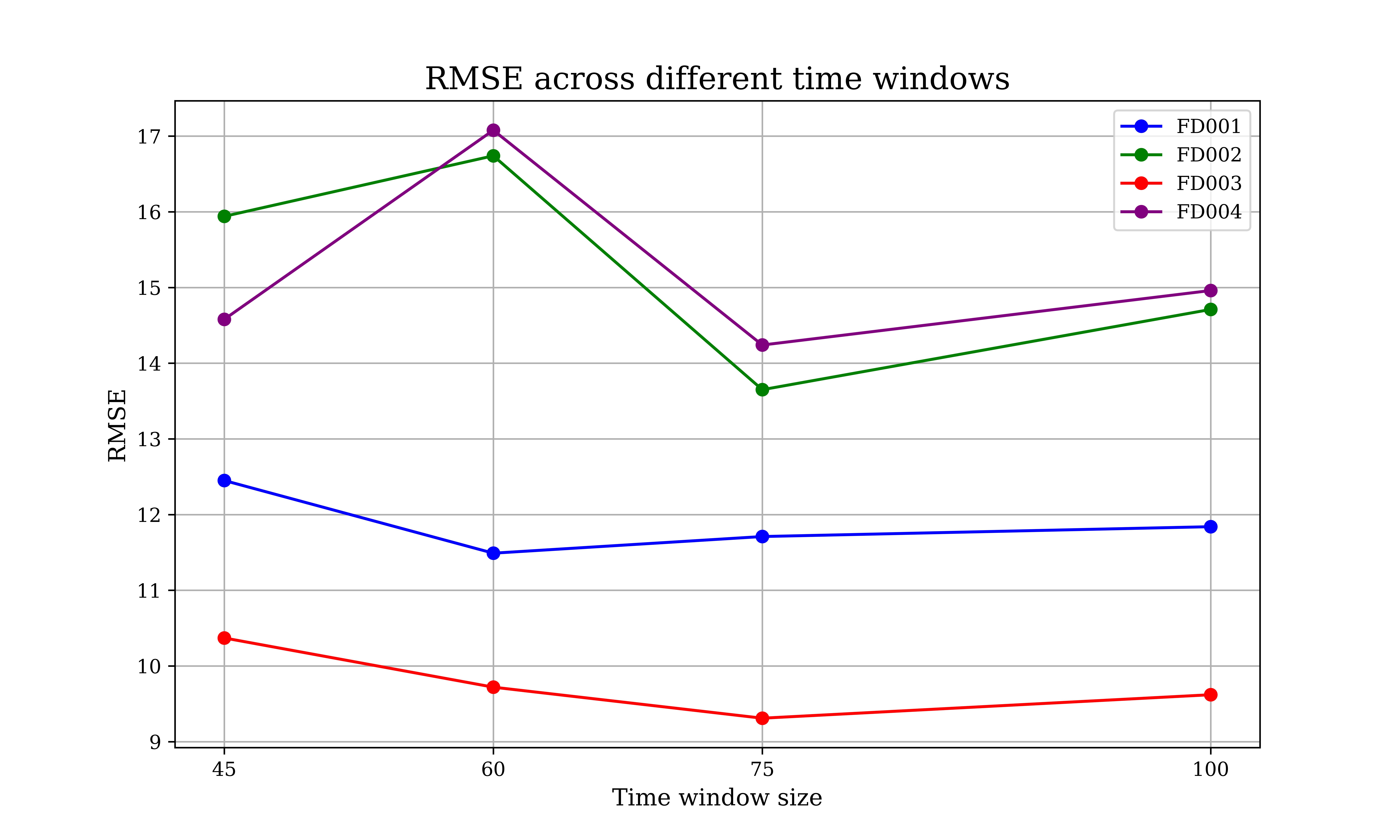}
		\caption{RUL prediction's RMSE at different medium time-window size=45, 60, 75, and 100.}
		\label{diffrent-size}
	\end{figure}

\begin{figure}
		\centering
		\includegraphics[width=0.75\linewidth]{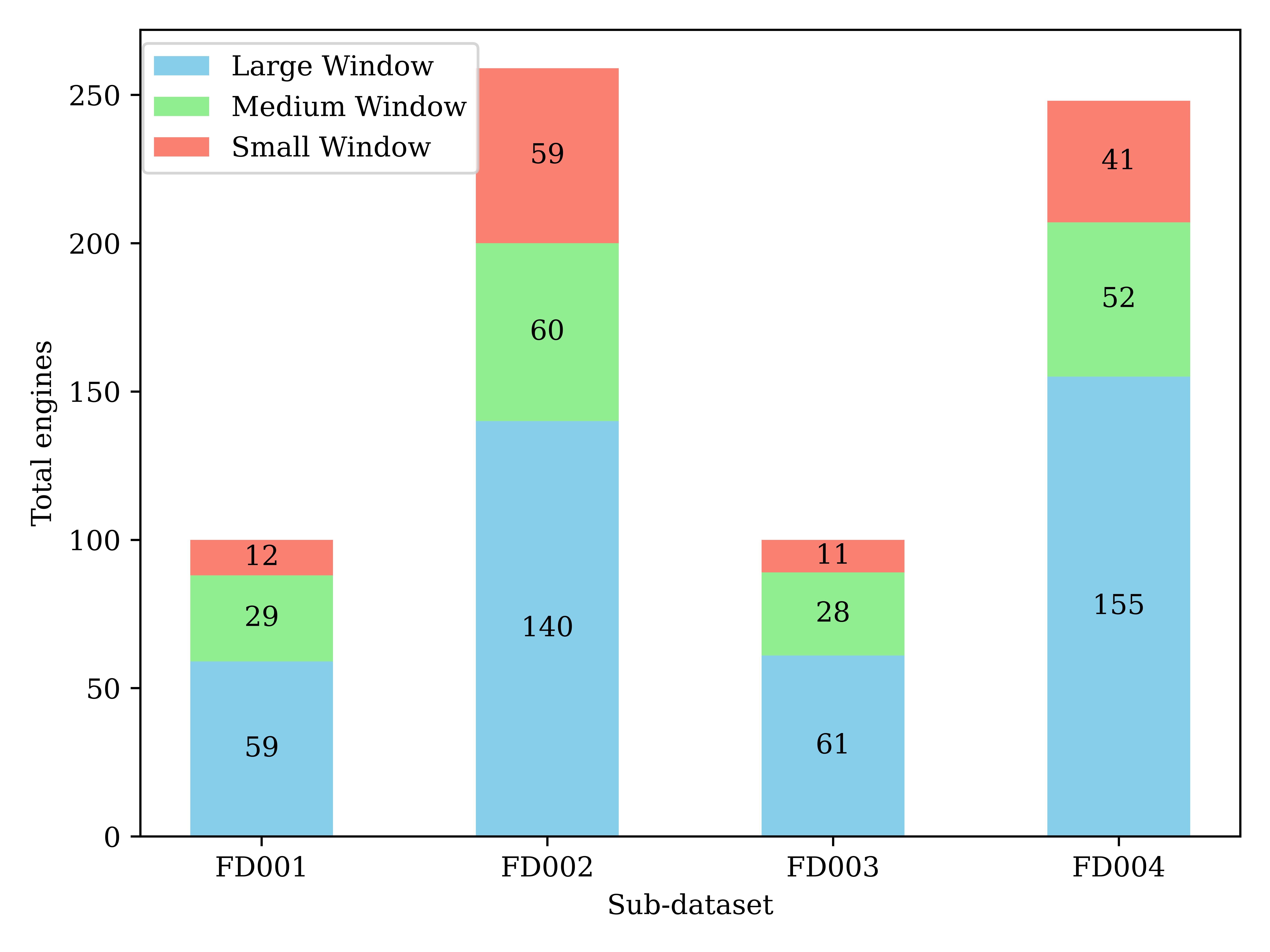}
		\caption{Engine counts by time window size for C-MAPSS sub-datasets.}
		\label{multitimefig}
	\end{figure}

	The proposed method employs multi-time windows to process the data. To cover every engine in the data, the smallest selected time window needs to be less than the least number of available engine sequences. This limitation is more pronounced in the test set than in the training data, because in the training set, the fault grows in magnitude until system failure, whereas in the test set, the time series ends prior to system failure \cite{4711414}. Consequently, the minimum number of available sequences in the test data influences the selection of time window sizes. Among the available time window sizes for the small one, we selected the longest time step for each sub-dataset. Thus, the small time window size for FD001, FD002, FD003, and FD004 was set to 31, 21, 38, and 19, due to its sequences limitation the same as the fixed time window size selected in \cite{Li2022}. To employ a wider view of historical data, the large time window size for all sub-datasets was chosen to be the same, set at 125. We implemented the medium-time window size after testing on various time window sizes. As shown in Fig. \ref{diffrent-size}, The lowest RMSE for FD001 was observed with a medium time-window size of 60, whereas for FD002, FD003, and FD004, it was achieved with a medium time-window size of 75. Fig. \ref{multitimefig} provides more details about implementing the approach in each sub-dataset and the number of engines utilized for each time-window size.

	\subsection{Experimental Results}
	
	\subsubsection{Model's performance}
	
	In this section, the model's performance in different time-window sizes is evaluated, and the results are reported for each time-window size in Table \ref{windowtable}.There are different number of engines covered with each time-window size, making RMSE the main metric for the comparison between different time-window sizes' results. It can be concluded from the table that longer sequences in large time window sizes led to lower RMSE than the medium and small time window sizes. However, it is not valid for comparison between medium and small sizes, and in FD001, FD002, and FD004, the small time-window size engines resulted in lower RMSE than the medium one.

		\begin{table*}[htbp]
		\centering
		\caption{Performance of TCFT-BED on each time window size.}
		\label{windowtable}
		\begin{tabular}{lcccccccc}
			\toprule
			\multicolumn{9}{c}{Time Window Size} \\ \cmidrule(lr){2-9}
			Sub-dataset & \multicolumn{2}{c}{Large} & \multicolumn{2}{c}{Medium} & \multicolumn{2}{c}{Small} & \multicolumn{2}{c}{Concatenation} \\
			\cmidrule(lr){2-3} \cmidrule(lr){4-5} \cmidrule(lr){6-7} \cmidrule(lr){8-9}
			& RMSE & Score & RMSE & Score & RMSE & Score & RMSE & Score \\
			\midrule
			FD001 & 8.00  & 57.07  & 16.57 & 182.95 & 9.51  & 21.24  & 11.53 & 261.26 \\
			FD002 & 9.33  & 184.88 & 18.96 & 468.31 & 15.59 & 484.13 & 13.65 & 1137.32\\
			FD003 & 5.64  & 32.67  & 12.16 & 72.90  & 14.92 & 55.71  & 9.31  & 161.28 \\
			FD004 & 11.61 & 364.17 & 20.05 & 535.82 & 14.31 & 309.75 & 14.24 & 1209.45\\
			\bottomrule
		\end{tabular}%
	\end{table*}

	In order to calculate RMSE and score for the whole sub-dataset, the model is trained and tested with every time window size, and the results of each one are saved, respectively, until the whole test part of the sub-dataset is evaluated. In other words, it saves results from a time window size and then adds results from another time window size to the results from the last evaluation, which continues until every engine in the test data is evaluated. There is no overlap in test data time windows, and all of the engines are covered through the multi-time window approach. Thus, adding the results from different time-window sizes is reasonable and aligns well with the main approach.
	
\begin{figure}[H]
		\centering
		\includegraphics[width=0.7\linewidth]{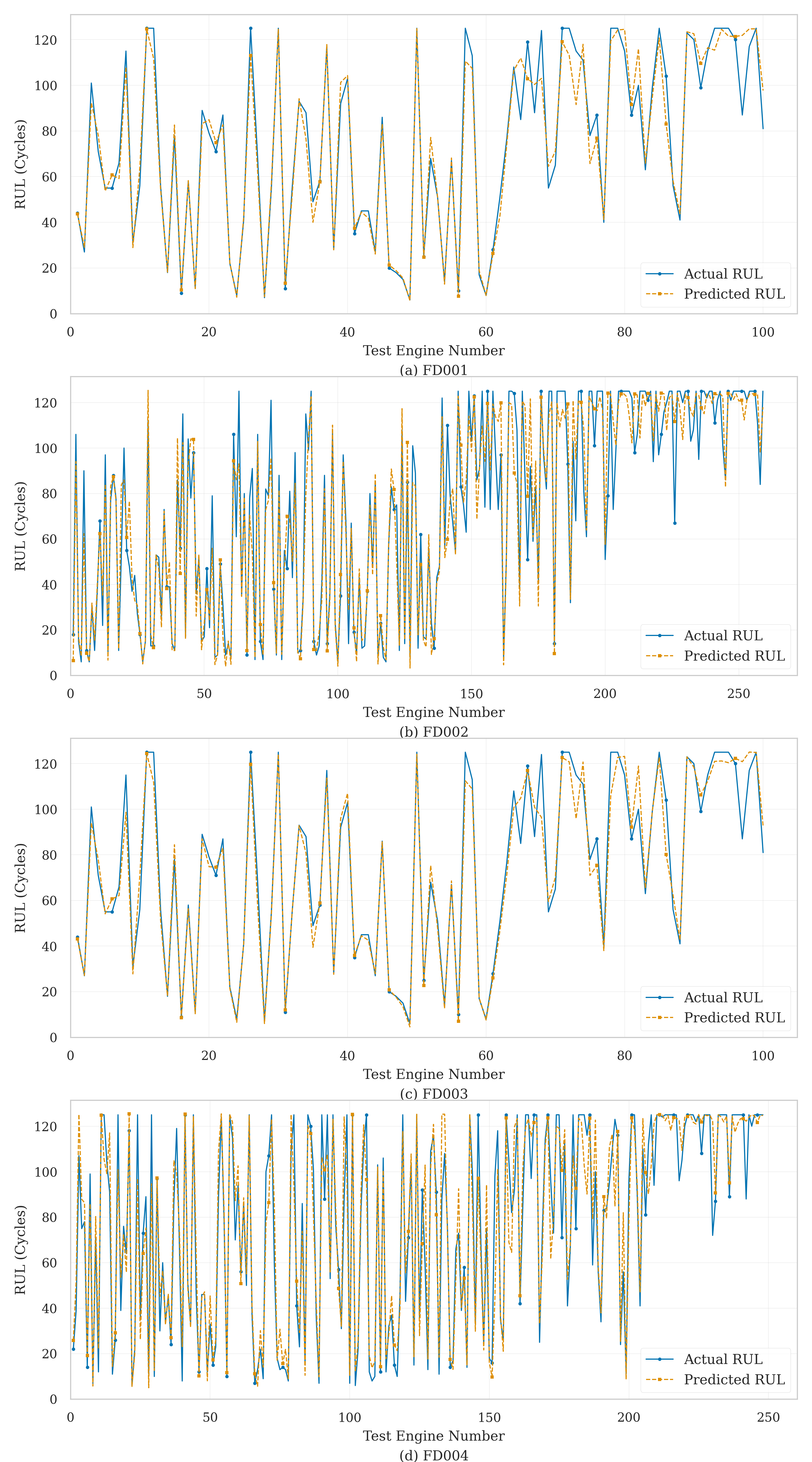}
		\caption{Prediction results with targets of total engines for each sub-data set of C-MAPSS.}
		\label{predsvs}
	\end{figure}

	To provide a visual representation of model's overall performance, the predicted and the real RULs of all the engines of the dataset are illustrated in Fig. \ref{predsvs}. Clearly, the model has shown remarkable accuracy in prediction, specially in FD001 and FD003 cases that have less fault modes and operational conditions than FD002 and FD004. it also highlights the prediction's high fluctuations over the targets in RULs between about 60 to 100 in all sub-datasets, where the medium time-window size plays its critical role.

\begin{figure*}
		\centering
		\includegraphics[width=0.99\linewidth]{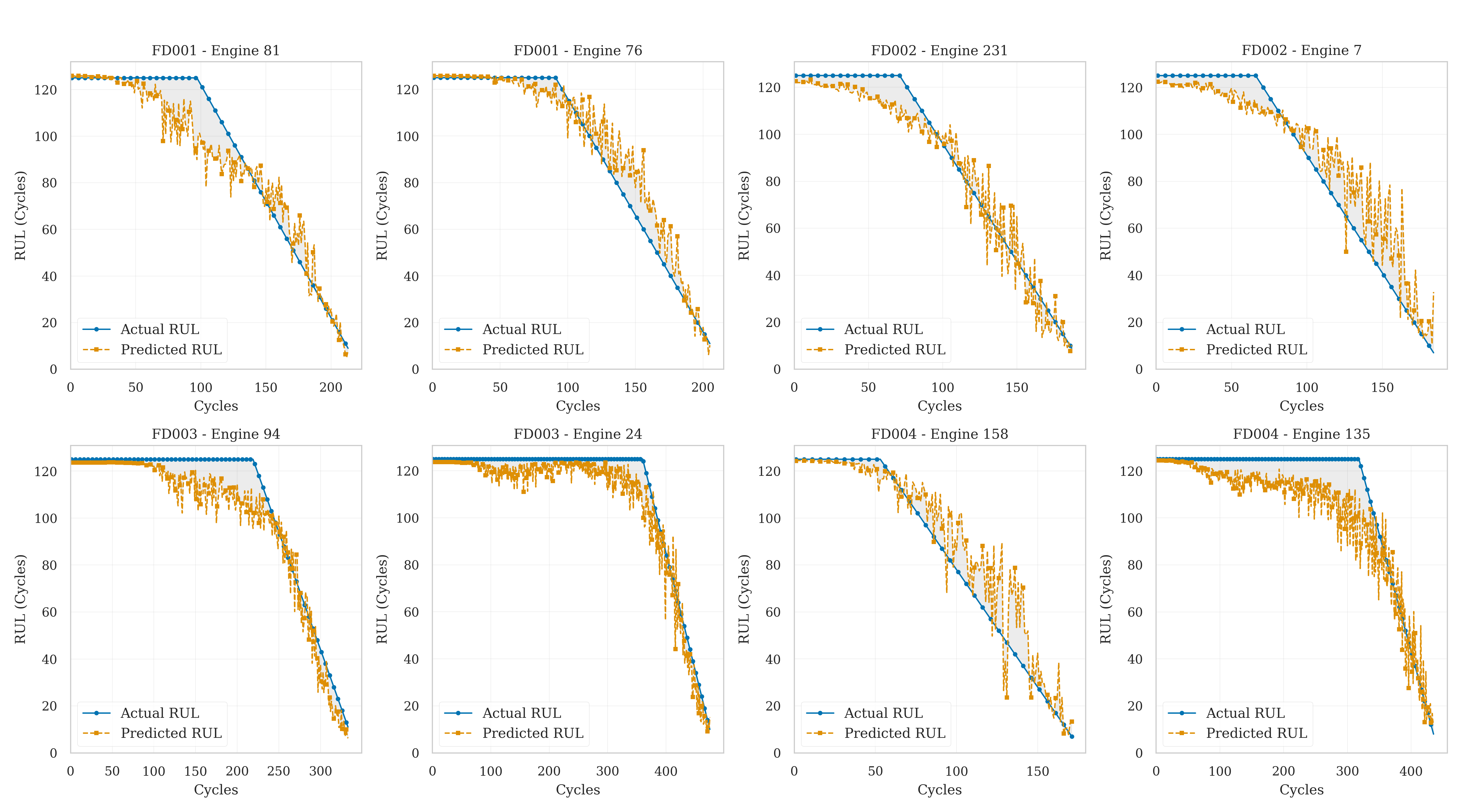}
		\caption{Visual validation of the model with lifetime predictions alongside real targets for two random engines from each sub-dataset: engines 81st and 76th of FD001, engines 231st and 7th of FD002, engines 94th and 24th of FD003, and engines 158th and 135th of FD004.}
		\label{predsvsengines}
	\end{figure*}

	In Fig. \ref{predsvsengines}, the lifetime RUL prediction of two random engines of each sub-dataset is depicted which provides visualizations about the prediction consistency over the lifetime targets.
	It clearly illustrates that the predictions closely align with the actual targets and progressively converge, particularly as the lifetime nears its end.
	
\begin{figure*}
		\centering
		\includegraphics[width=0.99\linewidth]{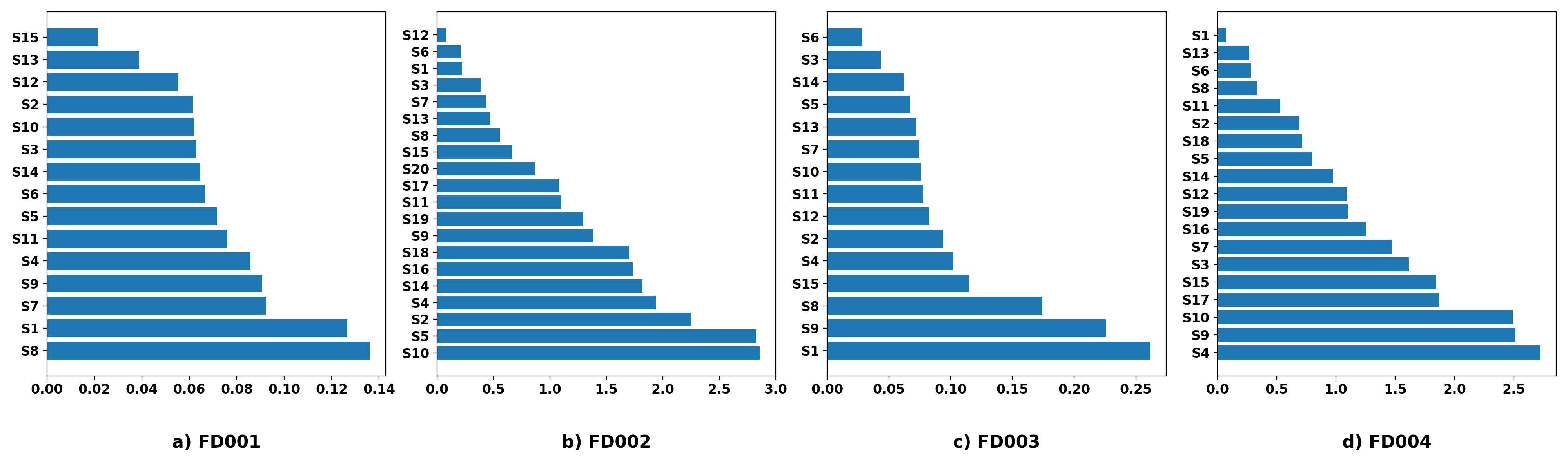}
		\caption{Global SHAP feature importance for the selected input sensors across the four C-MAPSS subsets (FD001–FD004). The bar lengths represent the mean absolute SHAP values, indicating each feature’s overall contribution to the RUL prediction. The most influential input channels vary across datasets, highlighting the model’s adaptability to different operational conditions and fault patterns.}
		\label{SHAP}
	\end{figure*}

	SHAP (SHapley Additive exPlanations) is used to interpret the model’s predictions by attributing importance values to each input feature based on its marginal contribution. Fig. \ref{SHAP} shows the global SHAP feature importance for the selected input features across the four C-MAPSS subsets (FD001–FD004). The top contributing sensors differ across subsets, reflecting the underlying complexity and operational conditions. For example, S8 and S1 are most influential in FD001, while FD004 relies more heavily on S4, S9, and S10. All selected input features are included in this summary, demonstrating how the model dynamically adapts feature usage depending on the dataset characteristics.
	
\begin{figure*}
		\centering
		\includegraphics[width=0.99\linewidth]{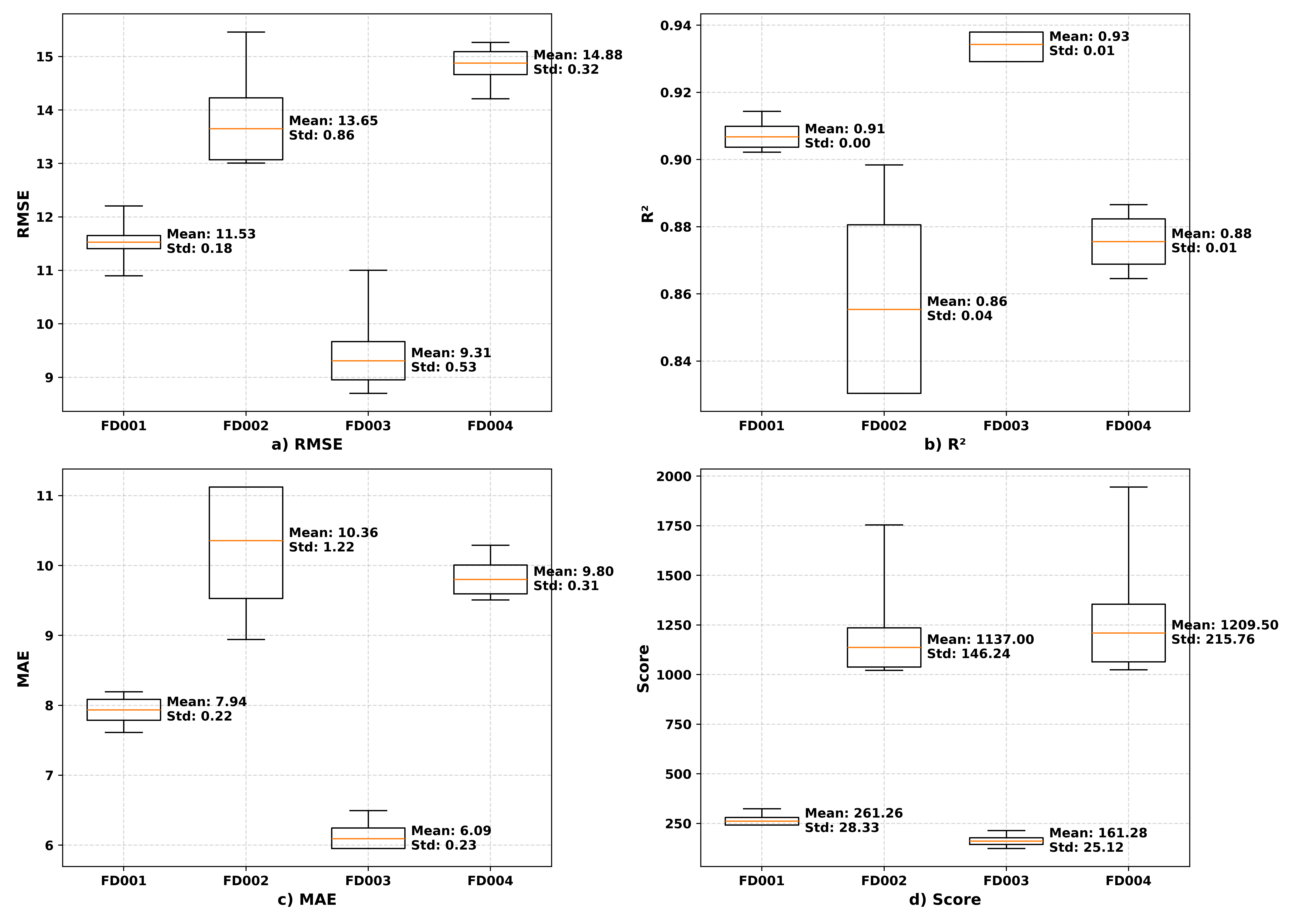}
		\caption{Box plots of model performance metrics (RMSE, R², MAE, and Score) across four sub-datasets (FD001, FD002, FD003, FD004).}
		\label{boxplot}
	\end{figure*}

	Fig. \ref{boxplot} shows model uncertainty across four sub-datasets (FD001--FD004) without manual parameter tuning, evaluated using RMSE, R\textsuperscript{2}, MAE, and Score. FD001 exhibits relatively low variability, with moderate RMSE (11.53~$\pm$~0.18) and MAE (7.94~$\pm$~0.22), and stable R\textsuperscript{2} (0.91~$\pm$~0.005). FD002 shows higher uncertainty, especially in MAE (10.36~$\pm$~1.22) and RMSE (13.65~$\pm$~0.86), while R\textsuperscript{2} (0.86~$\pm$~0.04) remains moderately stable. FD003 is the most consistent across all metrics, with the lowest standard deviations in RMSE (9.31~$\pm$~0.53), MAE (6.09~$\pm$~0.23), and R\textsuperscript{2} (0.93~$\pm$~0.01), indicating highly stable performance. FD004 displays the highest variability, particularly in Score (1209.5~$\pm$~215.76), suggesting less stable predictions likely due to the increased complexity of its operating conditions. Overall, FD003 demonstrates the most reliable performance, while FD002 and FD004 show greater model uncertainty.

	\subsubsection{Comparison with state-of-the-art methods}
	
	To demonstrate the superiority of the proposed TCFT-BED in prediction accuracy compared to state-of-the-art methods, we compared the results with those obtained from several advanced methods. These contrasting methods are relevant to our model or have demonstrated significant results. Conventional deep methods such as CNN \cite{li20181} and LSTM \cite{7998311} were included, alongside others like CNN-LSTM \cite{kong2019}, IDMFFN \cite{Li2022_IDMFFN}, MS-DCNN \cite{li2020}, CATA-TCN \cite{lin2024}, DA-TCN \cite{song2021}, 3D-AHNN \cite{10190349}, MTSTAN \cite{li2023}, and PSTFormer \cite{FU2025125995}, which are fusions of LSTMs, CNNs, or attention mechanisms with impressive accuracies in RUL prediction. Compared to these state-of-the-art methods, our model has achieved the lowest average RMSE and the second-best score across the entire dataset. The results of the proposed model with the mentioned methods for each sub-dataset and their RMSE and score average are provided in Table \ref{Results_comparison}, where the results are sorted based on the average RMSE.
	
	In detail, on FD003 and FD004, the proposed model has resulted in the best RMSE, and in FD002, it resulted in the second-best RMSE. Similar to the RMSE results, our score was not the best on FD001; however, we achieved the best performance on FD003 and the second-best result on FD004, demonstrating significant improvements.

	\begin{table*}[htbp]
		\centering
		\caption{Results comparison.}
		\label{Results_comparison}
		\resizebox{\textwidth}{!}{%
			\begin{tabular}{l c c c c c c c c c c c}
				\toprule
				\multirow{2}{*}{Model} & \multirow{2}{*}{Year} & \multicolumn{2}{c}{FD001} & \multicolumn{2}{c}{FD002} & \multicolumn{2}{c}{FD003} & \multicolumn{2}{c}{FD004} & \multicolumn{2}{c}{Average} \\
				\cmidrule(lr){3-4} \cmidrule(lr){5-6} \cmidrule(lr){7-8} \cmidrule(lr){9-10} \cmidrule(lr){11-12}
				& & RMSE & Score & RMSE & Score & RMSE & Score & RMSE & Score & RMSE & Score \\
				\midrule
				CNN & 2017 & 15.84 & 374.65 & 28.73 & 11394.12 & 13.53 & 318.93 & 30.66 & 14334.14 & 22.19 & 6605.46 \\
				LSTM & 2017 & 14.91 & 333.80 & 28.68 & 12908.59 & 13.37 & 316.55 & 30.32 & 11771.06 & 21.82 & 6332.50 \\
				CNN-LSTM & 2019 & 16.13 & 303.00 & 20.46 & 3440.00 & 17.12 & 1420.00 & 23.26 & 4630.00 & 19.24 & 2448.25 \\
				IDMFFN & 2022 & 12.18 & 204.69 & 19.17 & 1819.42 & 11.89 & 205.54 & 21.72 & 3338.84 & 16.24 & 1392.12 \\
				MS-DCNN & 2020 & \underline{11.44} & 196.22 & 19.35 & 3747.00 & 11.67 & 241.89 & 22.22 & 4844.00 & 16.17 & 2257.28 \\
				CATA-TCN & 2021 & 12.80 & 234.31 & 17.61 & 1361.23 & 13.16 & 290.63 & 21.04 & 2303.42 & 16.15 & 1047.40 \\
				3D-AHNN & 2023 & 13.12 & {231.01} & {13.93} & \textbf{759.84} & 12.15 & 195.56 & 20.24 & 1710.29 & 14.85 & {724.17} \\ 
				DA-TCN & 2021 & 11.78 & 229.48 & 16.95 & 1842.38 & 11.56 & 257.11 & {18.23} & 2317.32 & 14.63 & 1161.57 \\ 
				
				MTSTAN & 2023 & \textbf{10.97} & \textbf{175.36} & {16.81} & {1154.36} & \underline{10.90} & \underline{188.22} & 18.85 & {1446.20} & {14.38} & 776.32 \\ 
				PSTFormer & 2024 & 12.08 & \underline{224} & \textbf{13.00} & \underline{877} & 12.11 & 308 & \underline{14.38} & \textbf{1182} & \underline{12.89} & \textbf{650.25} \\ 
				\textbf{TCFT-BED (Proposed)} &  & 11.53 & {261.26} & \underline{13.65} & {1137.32} & \textbf{9.31} & \textbf{161.28} & \textbf{14.24} & \underline{1209.45} & \textbf{12.18} & \underline{692.33} \\ 
				\bottomrule
			\end{tabular}%
		}
		\begin{flushleft}
			The best results are \textbf{bolded}, and the second best results are \underline{underlined}.
		\end{flushleft}
	\end{table*}

\begin{figure}[h]
		\centering
		\includegraphics[trim={0 0.5cm 0 0},clip,width=0.75\linewidth]{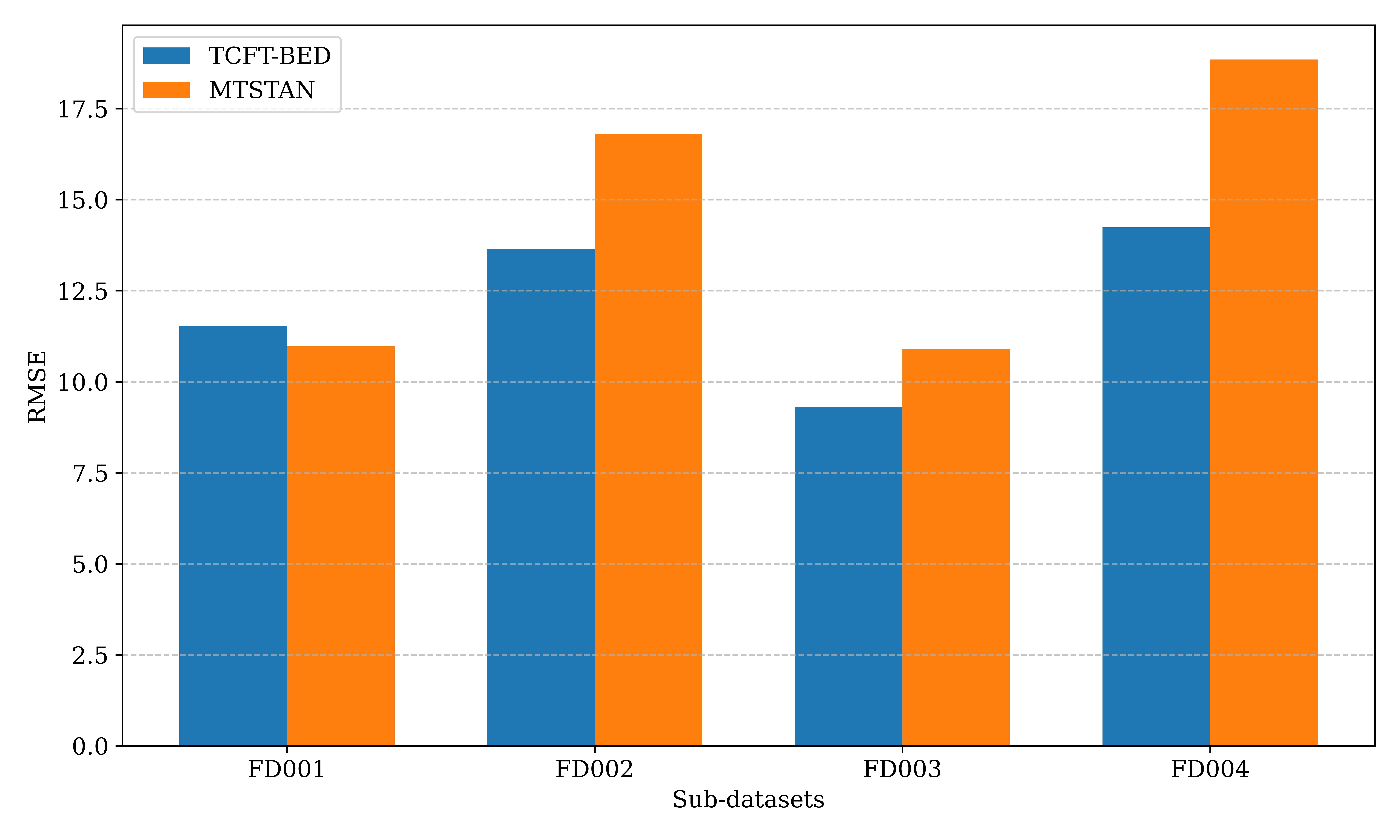}
		\caption{RMSE comparison between the proposed and MTSTAN as two state-of-the-art approaches with multi-time-window technique.}
		\label{compare-rmse-window}
	\end{figure}

	MTSTAN \cite{li2023}, as a multi-time-window approach similar to the proposed model, has also proved one of the best performances to time. This highlights the capabilities of the multi-time-window approach in RUL prediction. Fig. \ref{compare-rmse-window} compares the results of the proposed model alongside MTSTAN, identifying one of them as having the lowest average RMSE to date, while the other also demonstrates strong performance.
	
\begin{figure}[h]
		\centering
		\includegraphics[trim={0 0.5cm 0 0},clip,width=\linewidth]{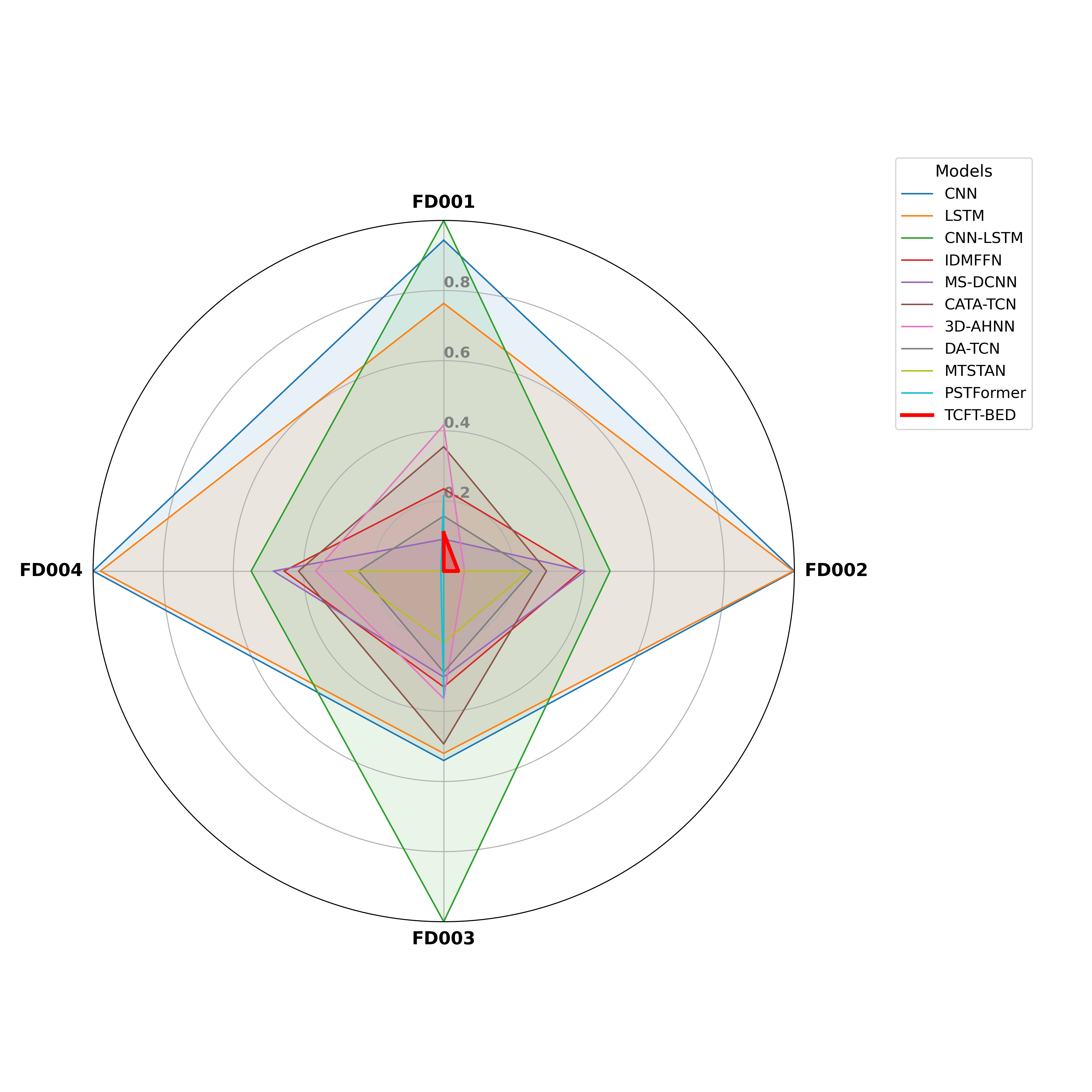}
		\caption{Radar chart of normalized RMSE across the four C-MAPSS subsets (FD001–FD004) for all compared models. Lower values (closer to the center) indicate better performance.}
		\label{Radar}
	\end{figure}

	Fig. \ref{Radar} shows a radar chart comparing the normalized RMSE performance of all evaluated models across the four C-MAPSS subsets (FD001–FD004). Each axis represents one dataset, and values are normalized between 0 and 1 per dataset for visual comparability. The proposed model, TCFT-BED, is highlighted in red and demonstrates consistently strong performance across all subsets, particularly excelling in FD003 and FD004.
	
\begin{figure}[h]
		\centering
		\includegraphics[trim={0 0.5cm 0 0},clip,width=\linewidth]{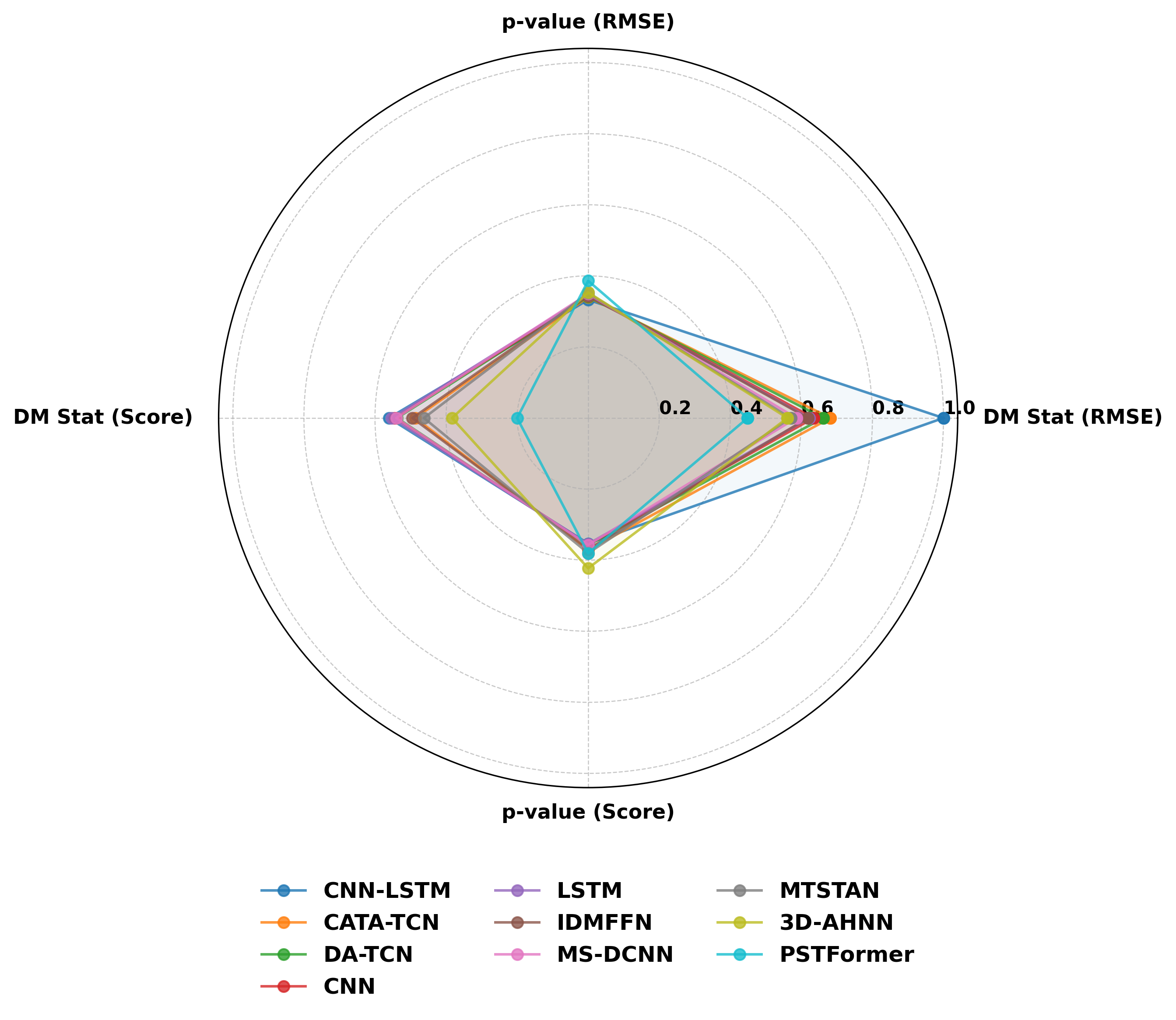}
		\caption{Radar plot of Diebold–Mariano (DM) statistics and p-values for ten predictive models based on average RMSE and Score.}
		\label{dieb}
	\end{figure}

	Fig. \ref{dieb} Radar plot of Diebold–Mariano (DM) statistics and p-values for ten predictive models across RMSE- and score-based evaluations. Based on the average RMSE and Score, models such as CNN-LSTM extend further on the DM Stat (RMSE) axis, indicating substantially less accurate predictions compared to the proposed method, whereas PSTFormer contracts toward the center on the DM Stat (Score) axis, reflecting predictions much closer to those of the proposed method. Mid-range models like DA-TCN lie between these extremes, showing moderate divergence but still outperforming less accurate baselines. Higher p-values near the outer rings denote weaker evidence against equal predictive accuracy, while lower p-values toward the center indicate stronger significance of performance differences.

	\subsubsection{Ablation study}
	In this section, we present a detailed ablation study to investigate the individual contributions of the critical components in the proposed TCFT-BED architecture. The goal is to isolate and evaluate the impact of each component on the model's overall prediction accuracy and robustness. We systematically remove or replace specific model components to perform this study and keep the remaining architecture and training parameters constant. Each ablation experiment is evaluated using the same training, validation, and test datasets of FD003. The performance is assessed using RMSE on the test set. The full hybrid model, incorporating TCN, Bi-LSTMs, and the gated attention block, serves as the baseline for comparison. The study is provided below:
	
	\begin{enumerate}
		\item \textbf{Configurating TCN Block} \\
		The first ablation experiment removes the TCN block from the model, and the Bi-LSTM and temporal gated attention components are used directly on the input features. This experiment assesses the importance of the TCN's ability to capture long-range dependencies via dilated convolutions and efficient time-series pattern extraction. After the removal, it was observed that without the TCN block, the model's capacity to capture high-level temporal patterns was reduced, leading to an increase in the prediction's RMSE by 15\%.
		
		We have also employed different dilation rates in the TCN block for each sub-dataset to choose a suitable dilation rate for each one. Table \ref{dilation} provides the results from the study and how it affects the prediction accuracy. It can be concluded from the table that FD001, FD003, and FD004 have shown remarkable performances with an exponential dilation rate of 1 whereas in FD002, the best performance resulted in a dilation rate of 8. Changing the dilation rate has affected the model's performance 
		
		\begin{table}[htbp]
			\centering
			\caption{Experimental results of each exponential dilation.}
			\label{dilation}
			\begin{tabular}{l c c c c}
				\toprule
				\multirow{2}{*}{Dilation} & \multicolumn{4}{c}{RMSE} \\
				\cmidrule(lr){2-5} 
				& FD001 & FD002 & FD003 & FD004 \\
				\hline
				1 & \textbf{11.53}& 13.79& \textbf{9.31}& \textbf{14.24}\\
				2 & 12.78& 15.31& 9.82& 14.54\\
				4 & 11.67& 15.51& 9.65& 14.49\\
				8 & 12.48& \textbf{13.65}& 10.21& 14.75\\
				16 & 11.75& 14.71& 9.50
				& 14.53\\
				\hline
			\end{tabular}
		\end{table}

		\item \textbf{Replacing Bi-LSTM Layers with standard LSTM layers} \\
		In the study, we implemented a model architecture consisting of Bi-LSTM encoder-decoder. To investigate this design choice's impact, we experimented with replacing each Bi-LSTM component with a standard LSTM network. This modification resulted in the creation of four distinct cases, the details of which are presented in Table \ref{case_lstm}. Additionally, Fig. \ref{caselstmfig} illustrates the superiority of the model in the best case and points out that although replacing both of the Bi-LSTMs with normal LSTMs as the key layers for sequence handling resulted in lower performance, it resulted in improved accuracy when the encoder was Bi-LSTM, and the decoder was the standard LSTM. 
		
		\begin{table}[htbp]
			\centering
			\caption{Experiment case setting for Bi-LSTMs.}
			\label{case_lstm}
			\begin{tabular}{l   c   c}
				\hline
				\textbf{Combination} & \textbf{Encoder} & \textbf{Decoder} \\
				\hline
				Case 1 & Bi-LSTM & Bi-LSTM \\
				
				Case 2 & Bi-LSTM & LSTM \\
				
				Case 3 & LSTM & Bi-LSTM \\
				
				Case 4 & LSTM & LSTM \\
				\hline
			\end{tabular}
		\end{table}
		
		\begin{figure}[h]
			\centering
			\includegraphics[width=0.75\linewidth]{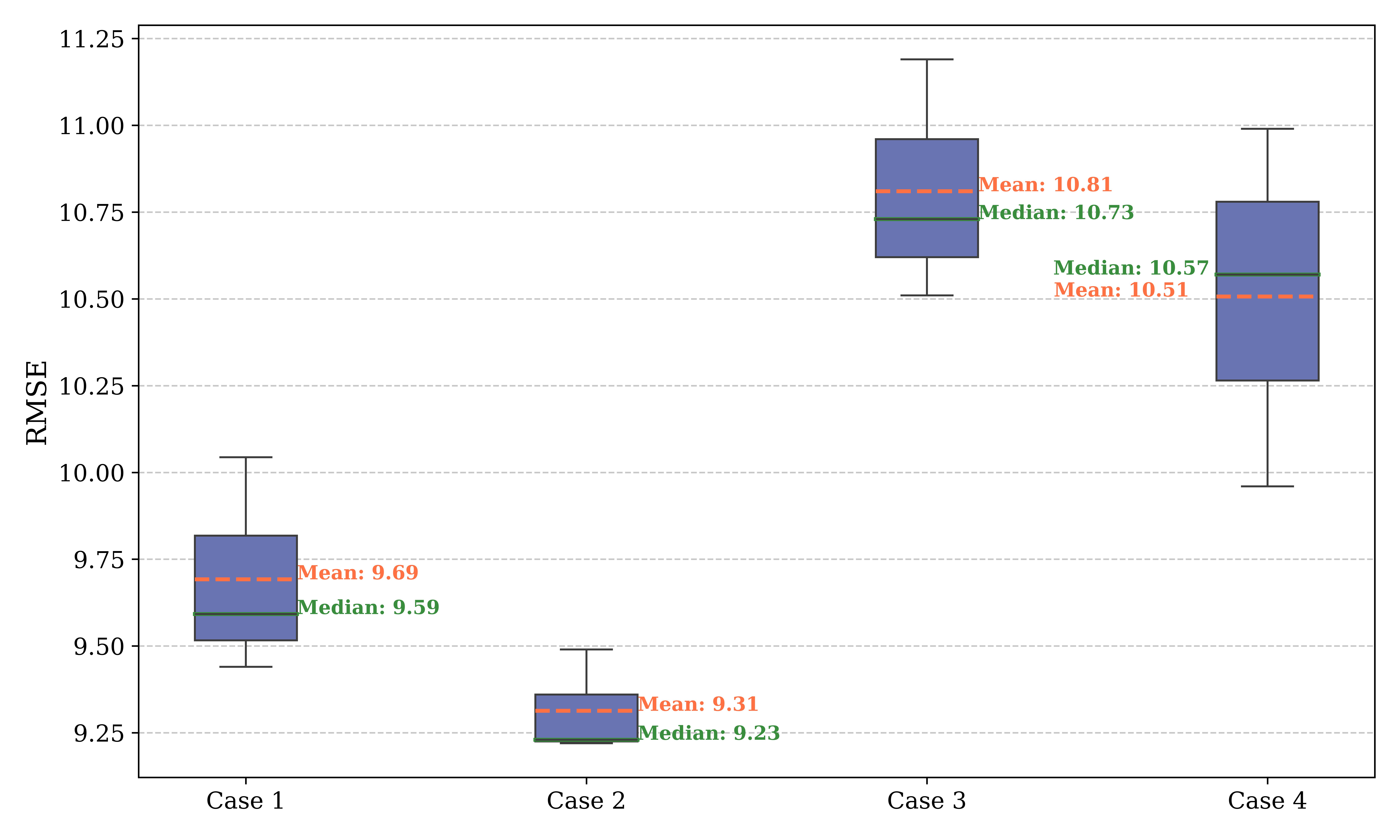}
			\caption{Boxplot of experimental results for four different cases.}
			\label{caselstmfig}
		\end{figure}

		\item \textbf{Replacing the gated attention block} \\
		For this ablation, we replace the gated attention block with a simple feed-forward network after the Bi-LSTM encoder layer, maintaining the exact input dimensions. This experiment examines the role of the attention mechanism and gating functionality within the gated attention block for handling temporal dependencies. Removing the attention and gating mechanisms significantly impacts performance and compared to the base model, the replaced model resulted in an 8\% higher RMSE because the model lost its ability to focus on critical time steps and features.
		\item \textbf{Comparison with Standard TFT and TCN Variants} \\
		To further validate the effectiveness of the proposed TCFT-BED, we extended the ablation study to include three baselines: the standard TFT, which employs static enrichment and attention; a pure TCN that models temporal patterns using only convolutional operations; and a reduced TCN variant derived from our baseline that removes the TFT block and Bi-LSTM layers to test parameter efficiency. Together with TCFT-BED, these models allow us to isolate the impact of each architectural component.

        As shown in Table \ref{tab:ablation_results}, TCFT-BED consistently achieves the best or second-best RMSE and score across FD001–FD004, clearly outperforming the standard TFT (e.g., RMSE reduced from 16.76 to 11.53 on FD001 and from 20.04 to 13.65 on FD002). The pure TCN performs worst overall, highlighting its inability to capture long-term and multi-scale degradation dependencies, while the reduced TCN variant offers parameter efficiency but sacrifices accuracy. These results confirm that the gains of TCFT-BED arise from the synergistic integration of TCN with the modified TFT block—without static enrichment, yielding more robust and coherent RUL prediction than any individual baseline. In all cases, inference time remained approximately 0.01 seconds or lower per run, since the test sequences are short and the main cost comes from loading the saved model rather than computation.

\begin{table*}[htbp]
\centering
\caption{Ablation study results on the FD001--FD004 subsets of the C-MAPSS dataset. 
Lower values indicate better performance. Average number of parameters and training time per epoch are also reported.}
\label{tab:ablation_results}
\resizebox{\textwidth}{!}{%
\begin{tabular}{l c c c c c c c c c c}
\toprule
\multirow{2}{*}{Model} & \multicolumn{2}{c}{FD001} & \multicolumn{2}{c}{FD002} & \multicolumn{2}{c}{FD003} & \multicolumn{2}{c}{FD004} & \multirow{2}{*}{Parameters} & \multirow{2}{*}{Avg Train Time (s)} \\
\cmidrule(lr){2-3} \cmidrule(lr){4-5} \cmidrule(lr){6-7} \cmidrule(lr){8-9}
 & RMSE & Score & RMSE & Score & RMSE & Score & RMSE & Score &  &  \\
\midrule
\textbf{TCFT-BED} & \textbf{11.53} & \textbf{261.26} & \textbf{13.65} & \textbf{1137.32} & \textbf{9.31} & \textbf{161.28} & \textbf{14.24} & \textbf{1209.45} & 81{,}153 & 5.51 \\
Original TFT & 16.76 & \underline{455.23} & 20.04 & 2157.24 & \underline{10.23} & 191.35 & 15.52 & 1452.26 & 101{,}409 & 6.52 \\
TCN & 18.42 & 1746.00 & 19.49 & 2061.60 & 12.48 & 241.84 & 18.14 & 2414.03 & \underline{70{,}849} & \textbf{3.50} \\
Modified TFT & \underline{12.76} & 704.02 & \underline{15.24} & \underline{1742.15} & 10.78 & \underline{181.26} & \underline{14.88} & \underline{1421.00} & \textbf{52{,}225} & \underline{4.92} \\
\bottomrule
\end{tabular}%
}
\end{table*}

    \end{enumerate}

	The ablation results confirm the complementary contributions of the TCN, Bi-LSTM, modifications on the base TFT, and gated attention blocks to the model's overall performance. The TCN block effectively captures long-range dependencies, while the Bi-LSTM layers are crucial for sequence encoding in both directions. With attention and gating mechanisms, the gated attention block significantly improves interpretability and accuracy by highlighting important temporal patterns.
	
	\section{Conclusion}
	In this study, we proposed a novel framework for RUL prediction that effectively integrates multiple strategies to enhance model performance. The model made high predicting accuracy by using the selection of sensors through monotonicity and correlation matrices, and the modified TFT block. Moreover, the application of TCNs with different dilation rates and a multi-time-window approach provided higher flexibility for the system. Our model has outperformed the state-of-the-art methods and demonstrated the best average RMSE of the C-MAPSS dataset with great improvements to time.
	
	In future works, it would be beneficial to look at more profound interactions of LSTMs with attention mechanisms, especially with regard to their suitability for multi-time-window strategies to handle inconsistent data lengths accurately and develop modern transformers with the ability to demonstrate a great performance in time series, particularly RUL prediction. Thus, this could further advance the model's applicability in real-world predictive maintenance scenarios.

\bibliographystyle{unsrt}  
\bibliography{references}  






\end{document}